%% file: paper-plain.tex
\documentclass{article} 
\usepackage{fourier}
\usepackage{amsmath,amssymb,amsthm}
\usepackage{fullpage}
\usepackage{url}
\usepackage[numbers]{natbib}
\usepackage{color}
\usepackage{algorithm,algorithmic}
\usepackage[colorlinks=true,linkcolor=blue]{hyperref}
\usepackage{url, color}

\include{header-plain}
\include{multiminimax}

\begin{document}

\title{Online Nonparametric Regression with General Loss Functions}

\author{Alexander Rakhlin\\
University of Pennsylvania
\and
Karthik Sridharan\\
Cornell University\\
}

\maketitle

\begin{abstract}
	This paper establishes minimax rates for online regression with arbitrary classes of functions and general losses.\footnote{This paper builds upon the study of online regression with square loss, presented by the authors at the COLT 2014 conference.} We show that below a certain threshold for the complexity of the function class, the minimax rates depend on both the curvature of the loss function and the sequential complexities of the class. Above this threshold, the curvature of the loss does not affect the rates.  Furthermore, for the case of square loss, our results point to the interesting phenomenon: whenever sequential and i.i.d. empirical entropies match, the rates for statistical and online learning are the same. 
	
In addition to the study of minimax regret, we derive a generic forecaster that enjoys the established optimal rates. We also provide a recipe for designing online prediction algorithms that can be computationally efficient for certain problems. We illustrate the techniques by deriving existing and new forecasters for the case of finite experts and for online linear regression.
\end{abstract}

\input{intro}

\input{prelim}

\input{upper_bounds}

\input{lower_bounds}

\input{main_results}

\input{algos}

\input{relatedwork}

\appendix
\input{appendix}

\bibliographystyle{plain}
\section*{Acknowledgements}
We gratefully acknowledge the support of NSF under grants CAREER DMS-0954737 and CCF-1116928, as well as Dean's Research Fund.

\bibliography{paper}

\end{document}

%% file: header-plain.tex
\usepackage{amsmath,amsfonts,amssymb,amsthm}
\usepackage{color}

\newtheorem{theorem}{Theorem}
\newtheorem{lemma}[theorem]{Lemma}

\newtheorem{corollary}[theorem]{Corollary}

\newtheorem{proposition}[theorem]{Proposition}

\newtheorem{remark}{Remark}

\theoremstyle{definition}
\newtheorem{definition}{Definition}

\DeclareMathOperator{\Ex}{\vphantom{p}\mathbb{E}}
\let\inf\undef
\DeclareMathOperator*{\inf}{\vphantom{p}inf}
\let\sup\undef
\DeclareMathOperator*{\sup}{\vphantom{p}sup}

\newcommand{\mbf}[1]{\mathbf{#1}}

\newcommand{\mrm}[1]{\mathrm{#1}}

\newcommand{\pred}{\widehat{y}}
\newcommand{\predset}{\widehat{\mathcal{Y}}}

\newcommand{\norm}[1]{\left\|#1\right\|}

\newcommand{\sign}{\mrm{sign}}
\newcommand{\argmin}[1]{\underset{#1}{\mrm{argmin}} \ }

\newcommand{\reals}{\mathbb{R}}
\newcommand{\En}{\mathbb{E}}  
\newcommand{\Es}[2]{\Ex_{#1}\left[ #2 \right]} 

\newcommand{\ind}[1]{{\bf 1}\left\{#1\right\}}

\newcommand\s{\mathbf{s}}
\newcommand\w{\mathbf{w}}
\newcommand\x{\mathbf{x}}

\newcommand\z{\mathbf{z}}

\renewcommand\v{\mathbf{v}}

\newcommand\cN{\mathcal{N}}

\newcommand\X{\mathcal{X}}

\newcommand\Y{\mathcal{Y}}
\newcommand\Z{\mathcal{Z}}
\newcommand\F{\mathcal{F}}
\newcommand\G{\mathcal{G}}
\newcommand\cH{\mathcal{H}}

\newcommand\fat{\mathrm{fat}}

\newcommand\Rad{\mathfrak{R}}

\newcommand\Reg{\mbf{Reg}}

\newcommand\cP{\mathcal{P}}

\newcommand\loss{\boldsymbol{\ell}}

\newcommand{\Rel}[2]{\mbf{Rel}_{#1}\left(#2 \right)}

\def\deq{\triangleq}

\newcommand{\UpperDelta}{\boldsymbol{\overline{\Delta}}}
\newcommand{\LowerDelta}{\boldsymbol{\underline{\Delta}}}
\newcommand{\bDelta}{\boldsymbol{\Delta}}

\newcommand{\bmu}{\ensuremath{\boldsymbol{\mu}}}
\newcommand{\bta}{\ensuremath{\boldsymbol{\eta}}}

%% file: intro.tex
\section{Introduction}

We study the problem of predicting a real-valued sequence $y_1,\ldots,y_n$ in an on-line manner. At time $t=1,\ldots,n$, the forecaster receives side information in the form of an element $x_t$ of an abstract set $\X$. The forecaster then makes a prediction $\pred_t$ on the basis of the current observation $x_t$ and the data $\{(x_i,y_i)\}_{i=1}^{t-1}$ encountered thus far, and then observes the response $y_t$. 

Such a problem of sequence prediction is studied in the literature under two distinct settings: probabilistic and deterministic \cite{mf-up-98}. In the former setting, which falls within the purview of time series analysis, one posits a parametric form for the data-generating mechanism and estimates the model parameters based on past instances and input information in order to make the next prediction. In contrast, in the deterministic setting one assumes no such probabilistic mechanism. Instead, the goal is phrased as that of predicting as well as the best forecaster from a benchmark set of strategies. This latter setting---often termed \emph{prediction of individual sequences}, or \emph{online learning}---is the focus of the present paper.

We let the outcome $y_t$ and the prediction $\pred_t$ take values in $\Y\subseteq\reals$ and $\predset\subseteq\reals$, respectively. Formally, a deterministic prediction strategy is a mapping $(\X\times\Y)^{t-1}\times \X \to \predset$. We let the loss function $(\pred_t,y_t)\mapsto \loss(\pred_t,y_t)$ score the quality of the prediction on a single round. 

Assume that the time horizon $n\in {\mathbb Z}_+$, is known to the forecaster. The overall quality of the forecaster is then evaluated against the benchmark set of predictors, denoted as a class $\F$ of functions $\X\to\predset$. The cumulative \emph{regret} of the forecaster on the sequence $(x_1,y_1),\ldots,(x_n,y_n)$ is defined as
\begin{align}
	\label{eq:standard_regret}
	\sum_{t=1}^n \loss(\pred_t,y_t) - \inf_{f\in\F} \sum_{t=1}^n \loss(f(x_t),y_t).
\end{align} 
The forecaster aims to keep the difference in \eqref{eq:standard_regret} small for \emph{all} sequences $(x_1,y_1),\ldots,(x_n,y_n)$.

The comparison class $\F$ encodes the prior belief about the family of predictors one expects to perform well. If a forecasting strategy guarantees small regret for all sequences, and if $\F$ is a good model for the sequences observed in reality, then the forecasting strategy will also perform well in terms of its cumulative error. In fact, we can take $\F$ to be a class of solutions (that is, forecasting strategies) to a set of probabilistic sources one would obtain by positing a generative model of data. By doing so, we are modeling solutions to the prediction problem rather than modeling the data-generating mechanism. We refer to \cite{mf-up-98,rakhlin2013semi} for further discussions on this ``duality'' between the probabilistic and deterministic approaches.

To ensure that $\F$ captures the phenomenon of interest, we would like $\F$  to be large. However, increasing the ``size'' of $\F$ likely leads to larger regret, as the comparison term in \eqref{eq:standard_regret} becomes smaller. On the other hand, decreasing the ``size'' of $\F$ makes the regret minimization task easier, yet the prediction method is less likely to be successful in practice. This dichotomy is an analogue of the bias-variance tradeoff commonly studied in statistics. A contribution of this paper is an analysis of the growth of regret (with $n$) in terms of various notions of complexity of $\F$. The task was already accomplished in \cite{RakSriTew10} for the case of absolute loss $\loss(a,b)=|a-b|$. In the present paper we obtain optimal guarantees for convex Lipschitz losses under very general assumptions.

To give the reader a sense of the results of this paper, we state the following informal corollary. Let complexity of $\F$ be measured via \emph{sequential entropy} at scale $\beta$, to be defined below. (For the reader familiar with covering numbers, this is a sequential analogue---introduced in \cite{RakSriTew10}---of the classical  Koltchinskii-Pollard entropy).  

\begin{corollary}[Informal]
	\label{cor:informal}
	Suppose sequential entropy at scale $\beta$ behaves as $\mathcal{O}(\beta^{-p})$, $p>0$. Then optimal regret
	\begin{itemize}
		\item for prediction with absolute loss grows as $n^{1/2}$ if $p\in(0,2)$, and as $n^{1-1/p}$ for $p>2$;
		\item for prediction with square loss grows as $n^{1-2/(2+p)}$ if $p\in(0,2)$, and as $n^{1-1/p}$ for $p>2$.
	\end{itemize}
	Moreover, these rates have matching, sometimes modulo a logarithmic factor, lower bounds.
\end{corollary}
The first part of this corollary is established in \cite{RakSriTew10}. The second part requires new techniques that take advantage of the curvature of the loss function.  

In an attempt to entice the reader, let us discuss two conclusions that can be drawn from Corollary~\ref{cor:informal}. First, the rates of convergence match optimal rates for excess square loss in the realm of distribution-free Statistical Learning Theory with i.i.d. data, under the assumption on the behavior of empirical covering numbers \cite{RakSriTsy14}. Hence, in the absence of a gap between classical and sequential complexities (introduced later) \emph{the regression problems in the two seemingly different frameworks enjoy the same rates of convergence}. A deeper understanding of this phenomenon is of a great interest.

The second conclusion concerns the same optimal rate $n^{-1/p}$ for both square and absolute loss for ``rich'' classes ($p>2$). Informally, strong convexity of the loss does not affect the rate of convergence for such massive classes. A geometric explanation of this interesting phenomenon requires further investigation.

We finish this introduction with a note about the generality of the setting proposed so far. Suppose $\X=\cup_{t\leq n} \Y^t$, the space of all histories of $\Y$-valued outcomes. Denoting $x_t=(y_1,\ldots,y_{t-1})\deq y^{t-1}$, we may view each $f\in\F$ itself as a strategy that maps history $y^{t-1}$ to a prediction. Ensuring that $x_t$ is not arbitrary but consistent with history only makes the task of regret minimization easier; the analysis of this paper for this case follows along the same lines, but we omit the extra overhead of restrictions on $x_t$'s and instead refer the reader to \cite{HanRakSri13,rakhlin2013semi}.

The paper is organized as follows. Section~\ref{sec:prelim} introduces the notation and then presents a brief overview of sequential complexities. Upper and lower bounds on minimax regret are established in Sections~\ref{sec:upper} and \ref{sec:lower}. We calculate minimax rates for various examples in Section~\ref{sec:rates}. We then turn to the question of developing algorithms in Section~\ref{sec:relaxations}. We first show that an algorithm based on the Rademacher relaxation is admissible (see \citep{rakhlin2012relax}) and yields the rates derived in a non-constructive manner in the first part of the paper. 
We show that further relaxations in finite dimensional space lead to the famous Vovk-Azoury-Warmuth forecaster. We also derive a prediction method for finite class $\F$.

%% file: prelim.tex
\section{Preliminaries}
\label{sec:prelim}

\subsection{Assumptions and Definitions}
\label{sec:assumptions}

We assume that the set of outcomes $\Y$ is a bounded set, a restriction that can be removed by standard truncation arguments (see e.g. \citep{gerchinovitz2013adaptive}).  
Let $\X$ be some set of covariates, and let $\F$ be a class of functions $\X\to\predset$ for some $\predset\subseteq\reals$. Recall the protocol of the online prediction problem: On each round $t\in \{1,\ldots,n\}$, $x_t\in\X$ is revealed to the learner who subsequently makes a prediction $\pred_t\in\predset$. The response $y_t\in \Y$ is revealed after the prediction is made. 

The loss function $\loss(\cdot, y)$ is assumed to be convex. Let $\partial_a \loss(a,y)$ denote any element of the subdifferential set (with respect to first argument), and assume that 
$$\sup_{a\in\predset,y\in\Y}|\partial_a \loss(a,y)|\leq G <\infty.$$
We assume that for any distribution of $y$ supported on $\Y$, there is a minimizer of expected loss that is finite and belongs to $\predset$:
$$ \predset \cap \argmin{\pred\in\reals} \En\loss(\pred,y) \neq \emptyset.$$
Given a $y\in\Y$, the error of a linear expansion at $a$ to approximate function value at $b$ is denoted by
$$\Delta_{a,b}^y ~\deq~ \loss(b,y) - [\loss(a,y) +  \partial_a \loss(a,y) \cdot \left(b - a\right)].$$	
Let $\LowerDelta:(\predset-\predset)\to \reals_{\geq 0}$ be a function defined pointwise as
\begin{align}
	\label{eq:lower_delta_function}
	\LowerDelta(x) = \inf_{a,b\in\predset, y\in\Y ~\text{s.t.}~ b-a=x} \Delta_{a,b}^y,
\end{align}
a lower bound on the residual for any two values separated by $x$. For instance, an easy calculation shows that $\LowerDelta(x)=x^2$ for  $\loss(\pred,y)=(\pred-y)^2$.

\subsection{Minimax Formulation}

Unlike most previous approaches to the study of online regression, we do not start from an algorithm, but instead work directly with minimax regret. We will be able to extract a (not necessarily efficient) algorithm after obtaining upper bounds on the minimax value. Let us introduce the notation that makes the minimax regret definition more concise. We use $\multiminimax{\cdots}_{t=1}^n$ to denote an  interleaved application of the operators, repeated over  $t = 1\ldots n$ rounds. With this notation, the minimax regret of the online regression problem described earlier can be written as
\begin{align}
	\label{eq:def_value}
	V_n = \multiminimax{\sup_{x_t}\inf_{\pred_t}\sup_{y_t}}_{t=1}^n\left\{ \sum_{t=1}^n \loss(\pred_t,y_t) - \inf_{f \in \F} \sum_{t=1}^n \loss(f(x_t),y_t)  \right\}
\end{align}
where each $x_t$ ranges over $\X$,  $\pred_t$ ranges over $\predset$, and $y_t$ ranges over $\Y$. 
An upper bound on $V_n$ guarantees the existence of an algorithm (that is, a way to choose $\pred_t$'s) with at most that much regret against any sequence. A lower bound on $V_n$, in turn, guarantees the existence of a sequence on which no method can perform better than the given lower bound.

\subsection{Sequential Complexities}

One of the key tools in the study of estimators based on i.i.d. data is the symmetrization technique \cite{GinZin84}. By introducing Rademacher random variables, one can study the supremum of an empirical process conditionally on the data. Conditioning facilitates the introduction of sample-based complexities of a function class, such as an empirical covering number. For a class of bounded functions, the covering number with respect to the empirical metric is necessarily finite and leads to a correct control of the empirical process even if discretization of the function class in a data-independent manner is impossible. We will return to this point when comparing our approach with discretization-based methods.

In the online prediction scenario, symmetrization is more subtle and involves the notion of a binary tree. The binary tree is, in some sense, the smallest entity that captures the sequential nature of the problem. More precisely, a $\Z$-valued tree $\z$ of depth $n$ is a complete rooted binary tree with nodes labeled by elements of a set $\Z$. Equivalently, we think of $\z$ as $n$ labeling functions, where $\z_1$ is a constant label for the root, $\z_2(-1),\z_2(+1)\in \Z$ are the labels for the left and right children of the root, and so forth. Hence, for $\epsilon=(\epsilon_1,\ldots,\epsilon_n)\in \{\pm1\}^n$, $\z_t(\epsilon)=\z_t(\epsilon_1,\ldots, \epsilon_{t-1})\in \Z$ is the label of the node on the $t$-th level of the tree obtained by following the path $\epsilon$. For a function $g:\Z\to\reals$, $g(\z)$ is an $\reals$-valued tree with labeling functions $g\circ \z_t$ for level $t$ (or, in plain words, evaluation of $g$ on $\z$). 

We now define two tree-based complexity notions of a class of functions.
\begin{definition}[\citep{RakSriTew10}]
Sequential Rademacher complexity of a class $\F\subseteq \reals^\X$ on a given $\X$-valued tree $\x$ of depth $n$, as well as its supremum, are defined as
\begin{align}
	 \Rad_n(\F; \x) \deq  \En \sup_{f\in\F}\left[ \sum_{t=1}^{n} \epsilon_t f(\x_t(\epsilon)) \right], ~~~~\Rad_n(\F) \deq \sup_{\x}\Rad_n(\F; \x)
\end{align}
where the expectation is over a sequence of independent Rademacher random variables $\epsilon= (\epsilon_1,\ldots,\epsilon_n)$. 
\end{definition}
One may think of the functions $\x_1,\ldots,\x_n$ as a predictable process with respect to the dyadic filtration  $\{\sigma(\epsilon_1,\ldots,\epsilon_t)\}_{t\geq 1}$. The following notion of a $\beta$-cover quantifies complexity of the class $\F$ evaluated on the predictable process.

\begin{definition}[\cite{RakSriTew10}]
		A set $V$ of $\reals$-valued trees of depth $n$ forms a $\beta$-cover (with respect to the $\ell_q$ norm) of a function class $\F\subseteq \reals^\X$ on a given $\X$-valued tree $\x$ of depth $n$ if 
		$$\forall f\in\F, \forall \epsilon\in\{\pm1\}^n, \exists \v\in V ~~~~\mbox{s.t.}~~~~ \frac{1}{n}\sum_{t=1}^n |f(\x_t(\epsilon))-\v_t(\epsilon)|^q \leq \beta^q.$$
		A $\beta$-cover in the $\ell_\infty$ sense requires that $|f(\x_t(\epsilon))-\v_t(\epsilon)| \leq \beta$ for all $t\in[n]$. The size of the smallest $\beta$-cover is denoted by $\cN_q(\beta, \F, \x)$, and $\cN_q(\beta, \F, n) \deq \sup_{\x} \cN_q(\beta, \F, \x)$.
\end{definition}

We will refer to $\log \cN_q(\beta, \F, n)$ as \emph{sequential entropy} of $\F$. In particular, we will study the behavior of $V_n$ when sequential entropy grows polynomially\footnote{It is straightforward to allow constants in this definition, and we leave these details out for the sake of simplicity.} as the scale $\beta$ decreases:
\begin{align}
	\label{eq:entropy_growth}
	\log \cN_2(\beta, \F, n)\sim \beta^{-p}, ~~~~p>0.
\end{align}
We also consider the parametric ``$p=0$'' case when sequential covering itself behaves as  
\begin{align}
	\label{eq:vc_growth}
	\cN_2(\beta,\F,n) \sim \beta^{-d}
\end{align}
(e.g. linear regression in a bounded set in $\reals^d$). We remark that the $\ell_\infty$ cover is necessarily $n$-dependent, so the forms we assume for nonparametric and parametric cases, respectively, are
\begin{align}
	\log\cN_\infty(\beta,\F,n) \sim \beta^{-p} \log (n/\beta)  ~~\mbox{ or }~~  \cN_\infty(\beta,\F,n) \sim (n/\beta)^{d}
\end{align}

%% file: upper_bounds.tex
\section{Upper Bounds} 
\label{sec:upper}

The following theorem from \cite{RakSriTew10} shows the importance of sequential Rademacher complexity for prediction with absolute loss.
\begin{theorem}[\cite{RakSriTew10}]
	\label{thm:absloss}
Let $\Y=[-1,1]$, $\F=[-1,1]^\X$, and $\loss(\pred,y)=|\pred-y|$. It then holds that
$$\Rad_n(\F) \leq V_n \leq 2\Rad_n(\F).$$
\end{theorem}
Furthermore, an upper bound of $2G\Rad_n(\F)$ holds for any $G$-Lipschitz loss. We observe, however, that as soon as $\F$ contains two distinct functions, sequential Radmeacher complexity of $\F$ scales as $\Omega(n^{1/2})$. Yet, it is known that minimax regret for prediction with square loss grows slower than this rate. Therefore, the direct analysis based on sequential Rademacher complexity (and a contraction lemma) gives loose upper bounds on minimax regret. The key contribution of this paper is an introduction of an offset Rademacher complexity that captures the correct behavior.

In the next lemma, we show that minimax value of the sequential prediction problem with any convex Lipschitz loss function can be controlled via offset sequential Rademacher complexity. As before, let $\epsilon = (\epsilon_1,\ldots,\epsilon_n)$ where each $\epsilon_i$ is an independent Rademacher random variable.

\begin{lemma}
	\label{lem:value_upper_bound}
	Under the assumptions and definitions in Section~\ref{sec:assumptions}, the minimax rate is bounded by
	\begin{align}
		\label{eq:val_upper_0_alpha}
		V_n &\leq \sup_{\x,\bmu} \En \sup_{f\in\F}\left[ \sum_{t=1}^{n} 2G \epsilon_t (f(\x_t(\epsilon))-\bmu_t(\epsilon)) - \LowerDelta\Big(f(\x_t(\epsilon))-\bmu_t(\epsilon)\Big)  \right]
	\end{align}
	where $\x$ and $\bmu$ range over all $\X$-valued and $\predset$-valued trees of depth $n$, respectively. 
\end{lemma}
The right-hand side of \eqref{eq:val_upper_0_alpha} will be termed \emph{offset Rademacher complexity} of a function class $\F\subseteq \reals^\X$ with respect to a convex even \emph{offset function} $\LowerDelta:\reals\to \reals_{\geq 0}$ and a mean $\reals$-valued tree $\bmu$. If $\LowerDelta\equiv 0$, we recover the notion of sequential Rademacher complexity since $\En[\epsilon_t\bmu_t(\epsilon)]=0$.

A matching lower bound on the minimax value will be presented in Section~\ref{sec:lower}, and the two results warrant a further study of offset Rademacher complexity. To this end, a natural next question is whether the chaining technique can be employed to control the supremum of this modified stochastic process. As a point of comparison, we first recall that sequential Rademacher complexity of a class $\G$ of $[-1,1]$-valued functions on $\Z$ can be upper bounded via the Dudley integral-type bound
\begin{align}
	\label{eq:rad_by_dudley}
	\En \sup_{g\in\G}\left[ \sum_{t=1}^{n} \epsilon_t g(\z_t(\epsilon)) \right] \leq \inf_{\rho\in (0,1]} \left\{ 4\rho n + 12\sqrt{n}\int_{\rho}^1 \sqrt{\log\cN_2(\delta,\G,\z)}d\delta \right\}, 
\end{align}
for any $\Z$-valued tree $\z$ of depth $n$, as shown in \citep{RakSriTew14}. We aim to obtain tighter upper bounds on the offset Rademacher by taking advantage of the negative offset term.

To initiate the study of offset Rademacher complexity with functions $\LowerDelta$ other than quadratic, we recall the notion of a convex conjugate.
\begin{definition}
	For a convex function $\psi:\mathcal{D}\to \reals$ with domain $\mathcal{D}\subseteq\reals$, the convex conjugate $\psi^*:\reals\to\reals\cup\{+\infty\}$ is defined as 
	$$\psi^*(a) = \sup_{d\in \mathcal{D}} \left\{ ad - \psi(d)\right\}.$$
\end{definition}

The chaining technique for controlling a supremum of a stochastic process requires a statement about the behavior of the process over a finite collection. The next lemma provides such a statement for the offset Rademacher process.
\begin{lemma}\label{lem:genmass}
Let $\Delta$ be a convex, nonnegative, even function on $\reals$ and let    
$\Gamma^*$ denote the convex conjugate of the function $x\mapsto \Delta(\sqrt{|x|})$.
	Assume $\Gamma^*$ is nondecreasing. For any finite set $W$ of $\reals$-valued trees of depth $n$ and any constant $C>0$,
\begin{align}
\label{eq:finite_lemma_with_var}
& \En \max_{\w\in W}\left\{ \sum_{t=1}^{n} 2C \epsilon_t \w_t(\epsilon) -      \Delta\left(\w_t(\epsilon)\right)  \right\}  \le  \inf_{\lambda>0} \left\{
		\frac{1}{\lambda}\log\left| W\right| +   n\ \Gamma^*\left(2C^2 \lambda  \right)
	\right\}.
\end{align}
Further, for any $[-G,G]$-valued tree $\bta$,
\begin{align}
	\label{eq:finite_lemma_with_etas_no_var}
	\En \max_{\w\in W}\left[ \sum_{t=1}^{n} \epsilon_t \bta_t(\epsilon) \w_t(\epsilon) \right] \leq G\sqrt{2\log|W| \cdot \max_{\w\in W, \epsilon_{1:n}} \sum_{t=1}^n \w_n(\epsilon)^2} \ .
\end{align}
\end{lemma}
As an example, if $\Delta(x)=x^2$, an easy calculation shows that $\Gamma^*(1)=0$ and $\Gamma^*(y)=+\infty$ for any $y\neq 1$. Hence, the infimum in \eqref{eq:finite_lemma_with_var} is achieved at $\lambda = 1/(2C^2)$, and the upper bound becomes $2C^2 \log\left|W\right|$.

We can now employ the chaining technique to extend the control of the stochastic process beyond the finite collection.
\begin{lemma}
	\label{lem:general_decomp}
	Let $\Delta$ and $\Gamma^*$ be as in Lemma~\ref{lem:genmass}.
	For any $\Z$-valued tree $\z$ of depth $n$ and a class $\G$ of functions $\Z\to \reals$ and any constant $C>0$, 
	\begin{align*}
	\En \sup_{g\in\G}\left[ \sum_{t=1}^{n} 2 C \epsilon_t g(\z_t(\epsilon)) - \Delta\left(g(\z_t(\epsilon))\right) \right]  
	&\leq \inf_{\gamma > 0}\left\{ C\inf_{\rho\in(0,\gamma)} \left\{ 4 \rho n + 12 \sqrt{n}\int_{\rho}^\gamma \sqrt{\log\cN_\infty(\delta, \G,\z)}d\delta\right\} \right.  \\
	&\left.\hspace{1.3cm}+ \inf_{\lambda>0}\left\{\frac{1}{\lambda}\log \cN_\infty\left(\tfrac{\gamma}{2},\G,\z\right) +  n\ \Gamma^*\left(2 C^2 \lambda   \right) \right\} \right\}. 
	\end{align*}
\end{lemma}
\begin{remark}
	\label{rem:l2_no_log}
	For the case of $\Delta(x)=x^2$, it is possible to prove the upper bound of Lemma~\ref{lem:general_decomp} in terms of $\ell_2$ sequential covering numbers rather than $\ell_\infty$ (see \cite{RakSri14a}).
\end{remark}

Lemma~\ref{lem:general_decomp}, together with Lemma~\ref{lem:value_upper_bound}, yield upper bounds on minimax regret under assumptions on the growth of sequential entropy. Before detailing the rates, we present lower bounds on the minimax value in terms of the offset Rademacher complexity and combinatorial dimensions.

%% file: lower_bounds.tex
\section{Lower Bounds}
\label{sec:lower}

The function $\LowerDelta$, arising from uniform (or strong) convexity of the loss function, enters the upper bounds on minimax regret. For proving lower bounds, we consider the dual property, that of (restricted) smoothness. To this end, let $S \subseteq \predset$ be a subset satisfying the following condition: 
	\begin{align} \label{property_two_point} 
	\forall s\in S, ~\exists y_1(s),y_2(s) \in \Y ~~~~\text{ s.t. }~~~~ s\in\argmin{\pred\in\predset}  \frac{1}{2}\left(\loss(\pred,y_1(s))+\loss(\pred,y_2(s))\right). 	\end{align}
	For any such subset $S$, let $\UpperDelta_S:(\predset-S) \to \reals_{\geq 0}$ be defined as
	\begin{align}
		\label{eq:upper_delta_function}
		\UpperDelta_S (x) = \sup_{s\in S, b \in \predset \text{ s.t. } b-s=x}\max\left\{\Delta_{s,b}^{y_1(s)}, \Delta_{s,b}^{y_2(s)}\right\}.
	\end{align}
	We write $\UpperDelta_{\kappa}$ for the singleton set $S=\{\kappa\}$. 
	
	The lower bounds in this section will be constructed from symmetric distributions supported on two carefully chosen points. Crucially, we do not require a uniform notion of smoothness, but rather a condition on the loss that holds for a restricted subset $S$ and a two-point distribution. 
	
	As an example, consider square loss and $\Y=\predset=(-B,B)$. For any $s\in\predset$, we may choose the two points as $s\pm\delta \in \Y$, for small enough $\delta$, with the desired property. Then $S=\predset$ and $\UpperDelta_S (x) = x^2$.

\begin{lemma}
	\label{lem:lower_bound_two_points}
	Fix $R>0$. Suppose $S\neq \emptyset$ satisfies condition \eqref{property_two_point}, and suppose that for any $s\in S$,  
	$$\partial \loss(s,y_1(s)) = +R, ~~~\partial \loss(s,y_2(s)) = -R.$$
	Then for any $S$-valued tree $\bmu$ of depth $n$,
	\begin{align}\label{eq:lower_bd_rad_general}
		V_n  \ge \sup_{\x} \En \sup_{f\in\F} \left[ \sum_{t=1}^n \epsilon_t R \left(f(\x_t(\epsilon)) - \bmu_t(\epsilon)\right) - \UpperDelta_{\bmu_t(\epsilon)}(f(\x_t(\epsilon))-\bmu_t(\epsilon)) \right].
	\end{align}
\end{lemma}

The lower bound in \eqref{eq:lower_bd_rad_general} is an offset Rademacher complexity that matches the upper bound of Lemma~\ref{lem:value_upper_bound} up to constants, as long as functions $\LowerDelta$ and $\UpperDelta$  exhibit the same behavior. In particular, the upper and lower bounds match up to a constant for the case of square loss.

Our next step is to quantify the lower bound in terms of $n$ according to ``size'' of $\F$. In contrast to the more common statistical approaches based on covering numbers and Fano inequality, we turn to a notion of a combinatorial dimension as the main tool. 
\begin{definition}
	An $\X$-valued tree of depth $d$ is said to be $\beta$-shattered by $\F$ if there exists an $\reals$-valued tree $\s$ of depth $d$ such that
		$$\forall \epsilon\in\{\pm1\}^d, ~\exists f^\epsilon \in\F ~~~\mbox{s.t.}~~~ \epsilon_t (f^\epsilon(\x_t(\epsilon))-\s_t(\epsilon))\geq \beta/2 $$
		for all $t\in\{1,\ldots,d\}$. The tree $\s$ is called a \emph{witness}. The largest $d$ for which there exists a $\beta$-shattered $\X$-valued tree is called the (sequential) fat-shattering dimension, denoted by $\fat_\beta(\F)$.
\end{definition}
The reader will notice that the upper bound of Lemma~\ref{lem:general_decomp} is in terms of sequential entropies rather than combinatorial dimensions. The two notions, however, are closely related.
\begin{theorem}[\cite{RakSriTew14}] 
	\label{thm:cover_fat}
	Let $\F$ be a class of functions $\X\to[-1,1]$. For any $\beta>0$, 
	$$\cN_2(\beta, \F, n)\leq \cN_\infty(\beta, \F, n) \leq \left( \frac{2en}{\beta}\right)^{\fat_\beta(\F)}.$$
\end{theorem}

As a consequence of the above theorem, if $\log \cN_2(\beta, \F, n) \geq (c/\beta)^p$ and $\beta\geq 1/n$, then 
$\fat_\beta(\F)\geq (c'/\beta)^p/\log(n)$
where $c,c'$ may depend on the range of functions in $\F$.

The lower bounds will now be obtained assuming $\fat_\beta(\F) \geq \beta^{-p}$ behavior of the fat-shattering dimension, and the corresponding statements in terms of the sequential entropy growth will involve extra logarithmic factors, hidden in the $\widetilde{\Omega}(\cdot)$ notation.

\begin{lemma}
	\label{lem:lower_p_greater_2}
	Suppose the statement of Lemma~\ref{lem:lower_bound_two_points} holds for some $R>0$, and suppose 
	\begin{align}
		\label{eq:lipschitz_delta}
		\UpperDelta_{\bmu_t(\epsilon)}(\bmu_t(\epsilon) - f(\x_t(\epsilon))) \leq \frac{R}{2}|\bmu_t(\epsilon) - f(\x_t(\epsilon))|
	\end{align}
	for any $f\in\F$ and $\bmu,\x$ in the statement of Lemma~\ref{lem:lower_bound_two_points}. Then it holds that for 
	any $\beta>0$ and $n=\fat_\beta(\F)$,
	$$V_n \geq (R/2)n\beta.$$
	In particular, if $\fat_\beta(\F) \geq \beta^{-p}$ for $p>0$, we have
	$$\frac{1}{n} V_n \geq (R/2)n^{-1/p} \ .$$
\end{lemma}

As an example, consider the case of square loss with $\Y=[-B,B]$. Then we may take $S=\{0\}$, $y_1=B$, $y_2=-B$, and hence $R=2B$. We verify that \eqref{eq:lipschitz_delta} holds for $\predset=[-B/2,B/2]$.

\begin{lemma}
	\label{lem:lower_p_less}
	Suppose the statement of Lemma~\ref{lem:lower_bound_two_points} holds for some $R>0$. For any class $\F'$ and $\beta>0$, there exists a modified class $\F$ such that for all $\beta' < \beta$, $\fat_{\beta'}(\F') \le \fat_{\beta'}(\F)\leq 2\fat_{\beta'}(\F')+4$ and for $n>\fat_{\beta}(\F)$, 
	$$\frac{1}{n} V_n \geq \sup_{\beta, \kappa} \left\{\frac{R\ \beta}{2} \sqrt{\frac{\fat_\beta(\F) }{2 n}} -   \ \bDelta_\kappa\left(\frac{\beta}{2}  \right) \right\}.$$
\end{lemma}

Armed with the upper bounds of Section~\ref{sec:upper} and the lower bounds of Section~\ref{sec:lower}, we are ready to detail specific minimax rates of convergence for various classes of regression functions $\F$ and a range of loss functions $\loss$.

%% file: main_results.tex
\section{Minimax Rates}
\label{sec:rates}

Combining Lemma~\ref{lem:value_upper_bound} and Lemma~\ref{lem:general_decomp}, we can detail the behavior of minimax regret under an assumption about the growth rate of sequential entropy.
\begin{theorem}
	\label{thm:p_and_P}
	Let $r\geq 2, p>0$ and suppose the loss function and the function class are such that
	$$\LowerDelta(t) \ge K t^{r}, ~~~~ \log\cN_\infty(\beta,\F,n) \le \beta^{-p}\log (n/\beta).$$ 
	Then for $p\in(0,2)$, 
	\begin{align}
		\label{eq:upper_p_less_2}
	\frac{1}{n}V_n  \leq \min\left\{ c_{r,p}~ n^{-\frac{r}{2(r -1) + p}} G^{\frac{2r}{2(r-1)+p}}K^{-\frac{2 - p}{2(r -1) + p}}\log(n) ~,~~  c_\F G\log^{1/2}(n) n^{-1/2} \right\}.
	\end{align}
	and for $p>2$
	\begin{align}
		\label{eq:upper_p_more_2}
		\frac{1}{n}V_n  \leq c_p G\log^{1/2}(n) n^{-1/p}  
	\end{align}
	Here, $c_\F$ depends on $\sup_{f\in\F} |f|_\infty$. At $p=2$, the bound \eqref{eq:upper_p_more_2} gains an extra $\log(n)$ factor.
\end{theorem}

We match the above upper bounds with lower bounds under the assumption on the growth of the combinatorial dimension. 
\begin{theorem}
	\label{thm:lower_p_P}
Suppose the statement of Lemma~\ref{lem:lower_bound_two_points} holds for some $R>0$ and $\kappa\in S\subseteq \predset$. Let $r\geq 2$, $p\in(0,2)$, and assume
$$
\UpperDelta_\kappa\left(\beta/2\right) \le K \beta^{r}, ~~~~ \fat_\beta \ge  \beta^{-p}.
$$ 
Then there exists a function class such that for some constant $c_{p,r}>0$,
\begin{align*}
\frac{1}{n}V_n & \geq c_{p,r} \min\left\{ n^{-\frac{r}{2(r -1) + p }}R^{\frac{2r}{2(r-1)+p}}K^{-\frac{2-p}{ 2(r-1) + p}}~,~~  Rn^{-1/2} \right\}.
\end{align*}
for $p\in(0,2)$. Furthermore, for $p>2$, for any $\F$ with $\fat_\beta \ge  \beta^{-p}$, 
\begin{align*}
\frac{1}{n}V_n & \geq (R/2) n^{-1/p}
\end{align*}
under the assumption \eqref{eq:lipschitz_delta}.
\end{theorem}

The lower bound of Theorem~\ref{thm:lower_p_P} matches (up to polylogarithmic in $n$ factors) the upper bound of Theorem~\ref{thm:p_and_P} in its dependence on $n$, the dependence on the constant $K$, and in dependence on the size of the gradients $G$ (respectively, $R$). The rest of this section is devoted to the discussion of the derived upper and lower bounds for particular loss functions or particular classes of functions. 

\subsection{Absolute loss}
We verify that the general statements recover the correct rates for the case of $\loss(\pred, y)=|\pred-y|$. Since the absolute loss is not strongly convex, we take $K=0$ (and $\LowerDelta\equiv 0$). Theorem~\ref{thm:p_and_P} then yields the $\mathcal{\widetilde{O}}(n^{-1/2})$ rate for $p\in(0,2)$ and $\mathcal{\widetilde{O}}(n^{-1/p})$ for $p>2$, up to logarithmic factors. These rates are matched, again up to logarithmic factors, in Theorem~\ref{thm:lower_p_P}. Of course, the result already follows from Theorem~\ref{thm:absloss}. 

It is also instructive to check the case of $r\to\infty$. In this case, if $K$ is scaled properly by the range of function values, the function $\LowerDelta$ approaches the zero function, indicating absence of strong convexity of the loss. Examining the power $\frac{r}{2(r -1) + p }$ in Theorem~\ref{thm:p_and_P}, we see that it approaches $1/2$, matching the discussion of the preceding paragraph.

\subsection{Square loss}

The case of square loss $\loss(\pred,y)=(\pred-y)^2$ has been studied in \cite{RakSri14a}. In view of Remark~\ref{rem:l2_no_log}, we state the corollary below in terms of $\ell_2$ covering numbers, thus removing some logarithmic terms of Theorem~\ref{thm:p_and_P}. 
\begin{corollary}
	\label{cor:upper_l2}
	For a class $\F$ with sequential entropy growth $\log \cN_2(\beta, \F, n) \leq \beta^{-p}$,
	\begin{itemize}
		\item For $p> 2$, the minimax regret\footnote{For $p=2$, $\frac{1}{n}V_n \leq C\log(n)n^{-1/2}$.} is bounded as ~~$\frac{1}{n}V_n \leq Cn^{-1/p}$ 
		\item For $p\in (0,2)$, the minimax regret is bounded as ~~$\frac{1}{n}V_n \leq Cn^{-2/(2+p)}$
		\item For the parametric case \eqref{eq:vc_growth}, ~~$\frac{1}{n}V_n \leq C dn^{-1}\log(n)$
		\item For finite set $\F$,  ~~$\frac{1}{n}V_n \leq C n^{-1}\log|\F|$
	\end{itemize}
\end{corollary}
\begin{corollary}
	\label{cor:lower_l2}
	The upper bounds of Corollary~\ref{cor:upper_l2} are tight\footnote{The $\widetilde{\Omega}(\cdot)$ notation suppresses logarithmic factors}:
	\begin{itemize}
		\item For $p\geq 2$, for any class $\F$ of uniformly bounded functions with a lower bound of $\beta^{-p}$ on sequential entropy growth, $\frac{1}{n}V_n \geq \widetilde{\Omega}(n^{-1/p})$
		\item For $p\in(0,2]$, for any class $\F$ of uniformly bounded functions, there exists a slightly modified class $\F'$ with the same sequential entropy growth such that $\frac{1}{n}V_n \geq \widetilde{\Omega}(n^{-2/(2+p)})$
		\item There exists a class $\F$ with the covering number as in \eqref{eq:vc_growth},  such that $\frac{1}{n}V_n \geq \Omega(dn^{-1}\log(n))$
	\end{itemize}
\end{corollary}

\subsection{$q$-loss for $q \in (1,2)$}

Consider the case of $\ell(\hat{y},y) = |y - \hat{y}|^q$, for $q\in(1,2)$, which interpolates between the absolute value and square losses. 

\begin{corollary}
	\label{cor:lqloss_1_2}
	Suppose $\Y =\predset = [-1,1]$ and $\ell(\hat{y},y) = |y - \hat{y}|^q$ for $q\in(1,2)$. Assume complexity of $\F$ as in Theorems \ref{thm:p_and_P} and \ref{thm:lower_p_P} for some $p>0$. Then
	$$
	\frac{1}{n}V_n = \Theta\left(\min\left\{ \left(q-1\right)^{-\frac{2 - p}{2 + p}} n^{-\frac{2}{2 + p}} , n^{-1/2} \right\}\right)
	$$		
\end{corollary}

\subsection{$q$-loss for $q \ge 2$}
It is easy to check that for $q > 2$, $\loss(\cdot,y)=|\cdot-y|^q$ is $q$-uniformly convex, and thus
$$
\LowerDelta(t) \geq Ct^q
$$
The upper bound of 
$$
\frac{1}{n} V_n \le Cn^{-\frac{q}{2(q -1) + p}}  
$$
then follows from Theorem~\ref{thm:p_and_P}.

\subsection{Logistic loss}

The loss function $\loss(\hat{y},y) = \log(1+\exp\{-\hat{y} y\})$ is strongly convex and smooth if the sets $\Y,\predset$ are bounded. This can be seen by computing the second derivative with respect to the first argument:
$$\loss''(\hat{y},y) = y^2\frac{\exp\{\hat{y}y\}}{(1+\exp\{\hat{y}y\})^2}$$
We conclude that
	$$
	\frac{1}{n}V_n = \widetilde{\Theta}\left(\min\left\{ n^{-\frac{2}{2 + p}} , n^{-1/2} \right\}\right)
	$$	

Logistic loss is an example of a function with third derivative bounded by a multiple of the second derivative. Control of the remainder term in Taylor approximation for such functions is given in \cite[Lemma 1]{bach2010self}.
Other examples of strongly convex and smooth losses are the exponential loss and truncated quadratic loss. These enjoy the same minimax rate as given above. 

\subsection{Logarithmic loss}

The technique developed in this paper is not universal. In particular, it does not yield correct rates for rich classes of functions under the loss
$$\loss(\pred,y) = -\log(\pred) \ind{y=1} - \log(1-\pred) \ind{y=0}$$
for the problem of probability assignment and a binary alphabet $\Y=\{0,1\}$.
The suboptimality of Lemma~\ref{lem:value_upper_bound} is due to the exploding Lipschitz constant. However, a modified approach is possible, and will be carried out in a separate paper.

\subsection{Sparse linear predictors and square loss}

We now focus on quadratic loss and instead detail minimax rates for specific classes of functions. Consider the following parametric class. Let $\G =\{g_1,\ldots,g_M\}$ be  a set of $M$ functions such that each $g_i : \X \mapsto [-1,1]$. Define $\F$ to be the convex combination of at most $s$ out of these $M$ functions. That is 
$$
\F = \left\{ \sum_{j=1}^s \alpha_j g_{\sigma_j} : \sigma_{1:s} \subset [M], \forall j, \alpha_j \ge 0, \sum_{j=1}^s \alpha_j = 1 \right\}
$$
For this example note that the sequential covering number can be easily upper bounded: we can choose $s$ out of $M$ functions in ${M \choose s}$ ways and observe that pointwise metric entropy for convex combination of $s$ bounded functions at scale $\beta$ is bounded as $\beta^{-s}$. We conclude that
$$
\cN_2(\beta,\F,n) \le \left(\frac{eM}{s} \right)^s  \beta^{-s}
$$
From the main theorem, for the case of square loss, the upper bound is
$$
\tfrac{1}{n} V_n \le O\left(\frac{s \log(M/s) }{n}\right)~.
$$
The extension to other loss functions follows immediately from the general statements.

\subsection{Besov spaces and square loss}
	Let $\X$ be a compact subset of $\reals^d$. Let $\F$ be a ball in Besov space $B_{p,q}^s(\X)$. When $s>d/p$, pointwise metric entropy bounds at scale $\beta$ scales as $\Omega(\beta^{-d/s})$ \cite[p. 20]{Vovk06metric}. On the other hand, when $s < d/p$,  and $p > 2$, one can show that the space is a $p$-uniformly convex Banach space. From \citep{RakSriTew14}, it can be shown that sequential Rademacher can be upper bounded by $O(n^{1-1/p})$, yielding a bound on minimax rate. These two controls together give the bound on the minimax rate. The generic forecaster with Rademacher complexity as relaxation (see Section~\ref{sec:relaxations}), enjoys the best of both of these rates. More specifically, we may identify the following regimes:
	\begin{itemize}
		\item If $s\geq d/2$, the minimax rate is $\tfrac{1}{n} V_n\leq O\left(n^{-\frac{2s}{2s+d}}\right)$.
		\item If $s< d/2$, the minimax rate depends on the interaction of $p$ and $d,s$:
		\begin{itemize}
			\item if $p> \frac{d}{s}$, the minimax rate $\tfrac{1}{n} V_n \leq O\left(n^{- \frac{s}{d}}\right)$, ~ otherwise, the rate is $\tfrac{1}{n} V_n \leq O\left(n^{-\frac{1}{p}}\right)$
		\end{itemize}
	\end{itemize}

\subsection{Remarks: Experts, Mixability, and Discretization}
The problem of prediction with expert advice has been central in the online learning literature \cite{PLG}. One can phrase the experts problem in our setting by taking a finite class $\F=\{f^1,\ldots,f^N\}$ of functions. It is possible to ensure sublinear regret by following the ``advice'' $f^{I_t}(x_t)$ of a randomly chosen ``expert'' $I_t$ from an appropriate distribution over experts. The randomized approach, however, effectively linearizes the problem and does not take advantage of the curvature of the loss. The precise way in which the loss enters the picture has been investigated thoroughly by Vovk \cite{vovk1995game} (see also \cite{haussler1998sequential}). Vovk defines a mixability curve that parametrizes achievable regret of a form slightly different than \eqref{eq:standard_regret}. Specifically, Vovk allows a constant other than $1$ in front of the infimum in the regret definition. Such regret bounds are called ``inexact oracle inequalities'' in statistics. Audibert \cite{audibert2009fast} shows that the mixability condition on the loss function leads to a variance-type bound in his general PAC-based formulation, yet the analysis is restricted to the case of finite experts. While it is possible to repeat the analysis in the present paper with a constant other than $1$ in front of the comparator, this goes beyond the scope of the paper. Importantly, our techniques go beyond the finite case and can give correct regret bounds even if discretization to a finite set of experts yields vacuous bounds. 

	Let us emphasize the above point again by comparing the upper bound of Lemma~\ref{lem:general_decomp} to the bound we may obtain via a metric entropy approach, as in the work of \cite{Vovk06metric}. Assume that $\F$ is a compact subset of $C(\X)$ equipped with supremum norm. The metric entropy, denoted by $\cH(\epsilon,\F)$, is the logarithm of the smallest $\epsilon$-net with respect to the sup norm on $\X$. An aggregating procedure over the elements of the net gives an upper bound (omitting constants and logarithmic factors) 
	\begin{align}
		\label{eq:balance_pollard}
		n\epsilon + \cH(\epsilon, \F)
	\end{align}
on regret \eqref{eq:standard_regret}. Here, $n\epsilon$ is the amount we lose from restricting the attention to the $\epsilon$-net, and the second term appears from aggregation over a finite set. The balance \eqref{eq:balance_pollard} fails to capture the optimal behavior for large nonparametric sets of functions. Indeed, for an $O(\epsilon^{-p})$ behavior of metric entropy, Vovk concludes the rate of $O\left(n^{\frac{p}{p+1}}\right)$. For $p\leq 2$, this is slower than the $O\left(n^{\frac{p}{p+2}}\right)$ rate one obtains from Lemma~\ref{lem:general_decomp} by trivially upper bounding the sequential entropy by metric entropy. The gain is due to the chaining technique, a phenomenon well-known in statistical learning theory. Our contribution is to introduce the same concepts to the domain of online learning. 

%% file: algos.tex

\section{Relaxations and Algorithms}
\label{sec:relaxations}
To design generic forecasters for the problem of online non-parametric regression we follow the recipe provided in  \citep{rakhlin2012relax}. It was shown in that paper that if one can find a relaxation $\mbf{Rel}_n$ (a sequence of mappings from observed data to reals) that satisfies certain  conditions, then one can define prediction strategies based on such relaxations. Specifically, we look for relaxations that satisfy the initial condition 
\begin{align*}
\Rel{n}{x_{1:n},y_{1:n}} \ge - \inf_{f \in \F} \sum_{t=1}^n \loss(f(x_t),y_t)
\end{align*}
and the recursive admissibility condition that requires
\begin{align}\label{eq:admissibility}
\inf_{\pred_t} \sup_{y_t}\left\{ \loss(\pred_t, y_t) + \Rel{n}{x_{1:t},y_{1:t}} \right\}  \le \Rel{n}{x_{1:t-1},y_{1:t-1}}
\end{align}
for any $t \in [n]$ and any $x_t \in \X$. A relaxation $\mbf{Rel}_n$ satisfying these two conditions is said to be admissible, and it leads to an algorithm 
\begin{align}
	\label{eq:algo}
\pred_t =  \argmin{\pred\in\predset} \sup_{y_t\in\Y}\left\{ \loss(\pred,y_t) + \Rel{n}{x_{1:t},y_{1:t}} \right\}. 
\end{align}
For this forecast the associated bound on regret is 
\begin{align}
	\label{eq:reg_bound}
\Reg_n :=  \sum_{t=1}^n \loss(\pred_t,y_t) - \inf_{f\in\F} \sum_{t=1}^n \loss(f(x_t),y_t) \le \Rel{n}{\emptyset}
\end{align}
(see \citep{rakhlin2012relax} for details).  We now claim that the following conditional version of  \eqref{eq:val_upper_0_alpha} gives an admissible relaxation and leads to a method that enjoys the regret bounds shown in the first part of the paper.

\begin{lemma}\label{lem:condradrel}
The following relaxation is admissible:
\begin{align*}
\Rad_n(x_{1:t},y_{1:t}) = \sup_{\x,\bmu}\En_\epsilon \sup_{f\in\F}\left[\sum_{j=t+1}^{n} 2G \epsilon_j (f(\x_j(\epsilon))-\bmu_j(\epsilon)) - \LowerDelta\Big(f(\x_j(\epsilon))-\bmu_j(\epsilon)\Big) - \sum_{j=1}^t \ell(f(x_j) , y_j)  \right]. 
\end{align*}
The algorithm \eqref{eq:algo} with this relaxation enjoys the regret bound of offset Rademacher complexity
$$
\Reg_n \le \sup_{\x,\bmu} \En_\epsilon \sup_{f\in\F}\left[ \sum_{t=1}^{n} 2 G \epsilon_t  (f(\x_t(\epsilon))-\bmu_t(\epsilon)) - \LowerDelta\Big(f(\x_t(\epsilon)) - \bmu_t(\epsilon)\Big)  \right].
$$
\end{lemma}

The proof of Lemma~\ref{lem:condradrel} follows closely the proof of Lemma~\ref{lem:value_upper_bound} and we omit it (see \cite{rakhlin2012relax,StatNotes2012}). Since the regret bound for the above forecaster is exactly the one given in  \eqref{eq:val_upper_0_alpha}, the upper bounds in Corollary~\ref{cor:upper_l2} hold for the above algorithm. Therefore, the algorithm based on $\Rad_n(x_{1:t},y_{1:t})$ is optimal up to the tightness of the upper and lower bounds in Section~\ref{sec:lower} and Section~\ref{sec:upper}. 

For the rest of this section, we restrict our attention to the case when $\Y=\predset=[-B,B]$. We further assume that $\ell(\pred , y_t) + \Rel{n}{x_{1:t},(y_{1:t-1},y_t)}$ is a convex function of $y_t$. In this case, the prediction $\pred_t$ takes a simple form, as the supremum over $y_t$ is attained either at $B$ or $-B$. More precisely, the prediction can be written as
\begin{align}
	\label{eq:balance_minmax_quardatic}
\pred_t & = \argmin{\pred \in [-B,B]} \max\left\{\ell(\pred , B) + \Rel{n}{x_{1:t},(y_{1:t-1},B)}, \ell(\pred , - B) + \Rel{n}{x_{1:t},(y_{1:t-1},-B)}\right\}.
\end{align}

\subsection{Recipe for designing online regression algorithms for general loss functions}
We now provide a schema for deriving forecasters for general online non-parametric regression:

\begin{enumerate}
\item Find relaxation $\mbf{Rel}_n$ s.t. $
\Rad_n\left(x_{1:t},y_{1:t} \right) \le \Rel{n}{x_{1:t},y_{1:t}}$,\
and\ $\ell(\pred, y_t) + \Rel{n}{x_{1:t},y_{1:t-1},y_t}$ is convex in $y_t$.
\item Check the condition 
{\small
$$
\sup_{x_t \in \X, p_t \in \Delta([-B,B])}\left\{  \inf_{\pred_t}\Es{y_t \sim p_t}{\ell\left(\pred_t , y_t\right)} + \Es{y_t \sim p_t}{\Rel{n}{x_{1:t},y_{1:t}}} \right\} \le \Rel{n}{x_{1:t-1},y_{1:t-1}} 
$$}
\item Given $x_t$, the prediction $\pred_t$ is given by
$$
\pred_t  = \argmin{\pred \in [-B,B]} \max\left\{\ell(\pred , B) + \Rel{n}{x_{1:t},(y_{1:t-1},B)}, \ell(\pred , - B) + \Rel{n}{x_{1:t},(y_{1:t-1},-B)}\right\}
$$
\end{enumerate}

\begin{proposition}\label{prop:recipe}
For any algorithm derived from the above schema, $\Reg_n \le \Rel{n}{\emptyset}$.
\end{proposition}
The proof of this Proposition follows very closely the proof in \cite{rakhlin2012relax} (see also \cite{StatNotes2012}), and we omit it.

\subsubsection{Square Loss}
To provide concrete examples of how the recipe can be used to derive algorithms, we now consider the square loss setting, $\ell(\pred,y) = (\pred - y)^2$. In this case, we observe that in \eqref{eq:balance_minmax_quardatic}, the first term in the maximum decreases as $\pred$ increases to $B$ and likewise the second term monotonically decreases as $\pred$ decreases to $-B$. Hence, the solution to \eqref{eq:balance_minmax_quardatic} is given when both terms are equal (if this does not happen within the range $[-B,B]$ then we clip the prediction to this range). In other words, for the case of square loss, if we have an admissible relaxation, then the prediction based on this relaxation is simply given by:
$$
\pred_t = \mrm{Clip}\left(\frac{\Rel{n}{x_{1:t},(y_{1:t-1},B)} - \Rel{n}{x_{1:t},(y_{1:t-1},-B)} }{4B}\right)
$$ 
where $\mrm{Clip}(z)=B\ind{z>B}+(-B)\ind{z<-B}+z\ind{z\in[-B,B]}$. Hence, for any admissible relaxation such that $(\pred - y_t)^2 + \Rel{n}{x_{1:t},(y_{1:t-1},y_t)}$ is a convex function of $y_t$, the above prediction based on the relaxation enjoys the bound on regret $\mbf{Rel}_n$. Based on the above observations and the recipe outline, we now provide two examples of $\F$ for which algorithms are derived.\\

\noindent{\bf Example : Finite class of experts}\\
As an example of estimator derived from the schema for the square loss learning setting, we first consider the simple case $|\F| < \infty$. 

\begin{corollary}\label{cor:exp}
The following is an admissible relaxation:
$$
\Rel{n}{x_{1:t},y_{1:t}} = B^2 \log\left(\sum_{f\in\F} \exp\left(  - B^{-2} \sum_{j=1}^t (f(x_j) - y_j)^2\right) \right).
$$
It leads to the algorithm
\begin{align*}
\pred_t & =  \mrm{Clip}\left(\frac{B}{4} \log\left(\tfrac{\sum_{f\in\F} \exp\left(  - B^{-2} \sum_{j=1}^{t-1} (f(x_j) - y_j)^2 - B^{-2} (f(x_t) - B)^2\right)}{\sum_{f\in\F} \exp\left(  - B^{-2} \sum_{j=1}^{t-1} (f(x_j) - y_j)^2  - B^{-2} (f(x_t) + B)^2\right)} \right)  \right)
\end{align*}
which enjoys a regret bound
$
\Reg_n \le B^2 \log\left| \F \right| \ .
$
\end{corollary}

\noindent{\bf Example : Linear regression}\\
Next, consider the problem of online linear regression in $\reals^d$. Here $\F$ is the class of linear functions. For this problem we consider a slightly modified notion of regret,
$$
\sum_{t=1}^n (\pred_t - y_t)^2 - \inf_{f \in \F}\left\{ \sum_{t=1}^n (f^\top x_t - y_t)^2 + \lambda \norm{f}_2^2  \right\}.
$$
This regret can be seen alternatively as regret if we assume that on rounds $-d+1$ to $0$ Nature plays  $(\lambda e_1,0)$,\ldots, $(\lambda e_d,0)$, where $\{e_i\}$ are the standard basis vectors, and that on these rounds the learner (knowing this) predicts $0$, thus incurring zero loss over these initial rounds. We can readily apply the schema for designing an algorithm for this problem.

\begin{corollary}
	\label{cor:linreg}
For any $\lambda>0$, the following is an admissible relaxation:
{\small
\begin{align*}
\Rel{n}{x_{1:t},y_{1:t}} &= \norm{\sum_{j=1}^t y_j z_j}^2_{\left(\sum_{j=1}^t z_j z_j^\top + \lambda I\right)^{-1}} + 4 B^2 \log\left(\frac{\left(\frac{n}{d}\right)^d}{\Delta\left(\sum_{j=1}^t z_j z_j^\top + \lambda I\right)}\right)  - \sum_{j=1}^t y_j^2 \ .
\end{align*}}
It leads to the Vovk-Azoury-Warmuth forecaster \citep{Vovk98,AzouryWarmuth01}
\begin{align*}
\pred_t & = \mrm{Clip}\left( x_t^\top \left(\sum_{j=1}^t x_j x_j^\top + \lambda I\right)^{-1}  \left(\sum_{j=1}^{t-1} y_j x_j\right)
\right)
\end{align*}
and enjoys the regret bound
\begin{align*}
\frac{1}{n}\sum_{t=1}^n (\pred_t - y_t)^2 \le \frac{1}{n}  \sum_{t=1}^n (f^\top x_t - y_t)^2 + \frac{\lambda}{2 n} \norm{f}_2^2 + \frac{4 d B^2 \log\left(\frac{n}{\lambda d}\right)}{n}.
\end{align*}
\end{corollary}

The proofs of Corollaries~\ref{cor:exp} and \ref{cor:linreg} already appeared in \cite{RakSri14nonparametric}, and we omit them here.

%% file: relatedwork.tex
\section{Discussion and Related Work}

In the past twenty years, progress in online regression for arbitrary sequences, starting with the paper of  \cite{f-pwc-91}, has been almost exclusively on finite-dimensional \emph{linear} regression (an incomplete list includes \citep{Vovk98, haussler1998sequential, KivWar97, v-cos-01, cb-atgbaolr-99, acbg-ascola-02, AzouryWarmuth01, HazMeg07, gerchinovitz2013sparsity}). This is to be contrasted with Statistics, where regression has been studied for rich (nonparametric) classes of functions. Important exceptions to this limitation in the online regression framework -- and works that partly motivated the present findings -- are the papers of \cite{Vovk07wild,Vovk06metric,vovk2006hilbert}. Vovk considers regression with large classes, such as subsets of a Besov or Sobolev space, and remarks that there appears to be two distinct approaches to obtaining the upper bounds in online competitive regression. The first approach, which Vovk terms Defensive Forecasting, exploits uniform convexity of the space, while the second -- an aggregating technique (such as the Exponential Weights Algorithm) -- is based on the metric entropy of the space. Interestingly, the two seemingly different approaches yield distinct upper bounds, based on the respective properties of the space. In particular, Vovk asks whether there is a unified view of these techniques. The present paper addresses these questions and establishes optimal performance for online regression. 

Since most work in online learning is algorithmic, the boundaries of what can be proved are defined by the regret minimization algorithms one can find. One of the main algorithmic workhorses is the aggregating procedure (or, the Exponential Weights Algorithm). However, the difficulty in using an aggregating procedure beyond simple parametric classes (e.g. subsets of $\reals^d$) lies in the need for a ``pointwise'' cover of the set of functions -- that is, a data-independent cover in the supremum norm on the underlying space of covariates. The same difficulty arises when one uses PAC-Bayesian bounds \citep{audibert2009fast} that, at the end of the day, require a volumetric argument. Notably, this difficulty has been overcome in statistical learning, where it has long been recognized (since the work of Vapnik and Chervonenkis) that it is sufficient to consider an \emph{empirical} cover of the class -- a potentially much smaller quantity. Such an empirical entropy is necessarily finite, and its growth with $n$ is one of the key complexity measures for i.i.d. learning. In particular, the recent work of \cite{RakSriTsy14} shows that the behavior of empirical entropy characterizes the optimal rates for i.i.d. learning with square loss. To mimic this development, it appears that we need to understand empirical covering numbers in the sequential prediction framework.

A hint as to how to modify the analysis of \cite{RakSriTew10} for ``curved'' losses appears in the paper of \cite{cesa1999minimax} where the authors derived rates for log-loss via a two-level procedure: the set of densities is first partitioned into small balls of a critical radius $\gamma$; a minimax algorithm is employed on each of these small balls; and an overarching aggregating procedure combines these algorithms. Regret within each small ball is upper bounded by classical Dudley entropy integral (with respect to a pointwise metric) defined up to the $\gamma$ radius. The main technical difficulty in this paper is to prove a similar statement using ``empirical'' sequential covering numbers.

Interestingly, our results imply the same phase transition as the one exhibited in \citep{RakSriTew14} for i.i.d. learning with square loss. More precisely, under the assumption of the $O(\beta^{-p})$ behavior of sequential entropy, the minimax regret normalized by time horizon $n$ decays as $n^{-\frac{2}{2+p}}$ if $p\in(0,2]$, and as $n^{-1/p}$ for $p\geq 2$. We prove lower bounds that match up to a logarithmic factor, establishing that the phase transition is real. Even more surprisingly, it follows that, under a mild assumption that sequential Rademacher complexity of $\F$ behaves similarly to its i.i.d. cousin, \emph{the rates of minimax regret in online regression with arbitrary sequences match, up to a logarithmic factor, those in the i.i.d. setting of Statistical Learning}. This phenomenon has been noticed for some parametric classes by various authors (e.g. \citep{cbfhhsw-huea-97}). The phenomenon is even more striking given the simple fact that one may convert the regret statement, that holds for all sequences, into an i.i.d. guarantee. Thus, in particular, we recover the result of \citep{RakSriTsy14} through completely different techniques. Since in many situations, one obtains optimal rates for i.i.d. learning from a regret statement, the relaxation framework of \citep{rakhlin2012relax} provides a toolkit for developing improper learning algorithms in the i.i.d. scenario.

%% file: appendix.tex
\section{Proofs}

\begin{proof}[\textbf{Proof of Lemma~\ref{lem:value_upper_bound}}]
	Denoting the set of distributions on $\Y$ by $\cP$, minimax regret can be written as
	\begin{align}
		\label{eq:unrolled_minimax_with_swap}
		V_n &= \multiminimax{\sup_{x_t}\inf_{q_t}\sup_{p_t\in\cP}\underset{\underset{y_t \sim p_t}{\pred_t \sim q_t}}{\En}}_{t=1}^n \left\{\sum_{t=1}^n \loss(\pred_t,y_t) - \inf_{f\in\F}\sum_{t=1}^n \loss(f(x_t),y_t)  \right\}  \\
		&= \multiminimax{\sup_{x_t}\sup_{p_t\in\cP} \En_{y_t}}_{t=1}^n 
			\left\{
				\sum_{t=1}^n \inf_{\pred_t}  \En_{y_t}\left[\loss(\pred_t,y_t) \right] - \inf_{f\in\F}\sum_{t=1}^n \loss(f(x_t),y_t)  
			\right\} \notag \\
		&= \multiminimax{\sup_{x_t}\sup_{p_t\in\cP} \En_{y_t}}_{t=1}^n 
			\left[
				\sup_{f\in\F}
					\left\{
						\sum_{t=1}^n \inf_{\pred_t}\left\{ \En_{y_t}\left[\loss(\pred_t,y_t)\right] \right\} - \sum_{t=1}^n \loss(f(x_t),y_t)  
					\right\}
				\right] \notag
	\end{align}
	where the first equality is by definition, the second follows from an argument of \cite{AbeAgaBarRak09, RakSriTew10}, and the third is a simple rearrangement. Taking $\pred^*_t = \argmin{\pred \in \predset} \Es{y \sim p_t}{\loss(\pred,y)}$, we write the above as
	\begin{align}
		&\multiminimax{\sup_{x_t}\sup_{p_t\in\cP} \En_{y_t}}_{t=1}^n \left[\sup_{f\in\F}\left\{\sum_{t=1}^n \En_{y_t}\left[\loss(\pred^*_t,y_t) \right]  - \sum_{t=1}^n \loss(f(x_t),y_t)  \right\}\right] \label{eq:valbnd0}\\
			&= \multiminimax{\sup_{x_t}\sup_{p_t\in\cP} \En_{y_t}}_{t=1}^n\left[\sup_{f\in\F} \left\{\sum_{t=1}^n   \loss(\pred^*_t,y_t) -  \loss(f(x_t),y_t) \right\}\right] \label{eq:valbnd}
	\end{align}
	The last step holds true by observing that the terms $\loss(\pred^*_t,y_t)$ do not depend on $f$ and can therefore be moved outside the supremum over $f\in\F$. The equality then follows by the linearity of expectation. By  definition of $\LowerDelta$ in \eqref{eq:lower_delta_function},
	\begin{align}
		\label{eq:uppbd1}
	V_n &\le \multiminimax{\sup_{x_t}\sup_{p_t\in\cP} \En_{y_t}}_{t=1}^n\left[\sup_{f\in\F} \left\{\sum_{t=1}^n   \partial \loss(\pred_t^*,y_t) \cdot \left( \pred_t^* - f(x_t) \right) - \LowerDelta(\pred_t^* - f(x_t)) \right\}\right] 
	\end{align}
	By definition of $\pred^*_t$, we have, $\mathbb{E}_{y_t \sim p_t}\left[\partial \loss(\pred^*_t,y_t) \right] = \partial \mathbb{E}_{y_t \sim p_t}\left[\loss(\pred^*_t,y_t) \right] = 0$ by the assumption that the minimum is attained in $\predset$ (see Section~\ref{sec:assumptions}). Thus we can view $\left( \partial \loss(\pred_t^*,y_t) \right)_{t =1}^T$ as a martingale difference sequence. Hence, 
	\begin{align*}
	V_n 
		& \le \multiminimax{\sup_{x_t}\sup_{p_t\in\cP} \En_{y_t}}_{t=1}^n\Bigg[\sup_{f\in\F} \Bigg\{\sum_{t=1}^n   \left(\partial \loss(\pred_t^*,y_t) - \Es{y'_t \sim p_t}{\partial \loss(\pred_t^*,y'_t)} \right)\cdot \left( \pred_t^* - f(x_t) \right) - \LowerDelta(\pred_t^* - f(x_t)) \Bigg\}\Bigg]. 
	\end{align*}
	By Jensen's inequality the above is upper bounded by 
	\begin{align*}
		\multiminimax{\sup_{x_t}\sup_{p_t\in\cP} \En_{y_t,y'_t \sim p_t} \En_{\epsilon_t}}_{t=1}^n\Bigg[\sup_{f\in\F} \Bigg\{\sum_{t=1}^n   \epsilon_t \left(\partial \loss(\pred_t^*,y_t) - \partial \loss(\pred_t^*,y'_t) \right)\cdot \left( \pred_t^* - f(x_t) \right) - \LowerDelta(\pred_t^* - f(x_t))  \Bigg\}\Bigg] 
	\end{align*}
	where we introduced Rademacher random variables. The next step involves splitting the upper bound into two equal terms, one for the $y_t$ sequence and the other for the $y_t'$ sequence:
	\begin{align*}
		V_n&\le \multiminimax{\sup_{x_t}\sup_{p_t\in\cP} \En_{y_t \sim p_t} \En_{\epsilon_t}}_{t=1}^n\Bigg[\sup_{f\in\F} \Bigg\{\sum_{t=1}^n   2 \epsilon_t \partial \loss(\pred_t^*,y_t) \cdot \left( \pred_t^* - f(x_t) \right) - \LowerDelta(\pred_t^* - f(x_t))  \Bigg\}\Bigg] 
	\end{align*}
	Using Jensen's inequality once again leads to an upper bound of
	\begin{align*}
		&\multiminimax{\sup_{x_t}\sup_{p_t\in\cP} \En_{y_t \sim p_t} \En_{\epsilon_t}}_{t=1}^n\left[\sup_{f\in\F} \left\{\sum_{t=1}^n   2 \epsilon_t \partial \loss(\pred_t^*,y_t) \cdot \left( \pred_t^* - f(x_t) \right) - \LowerDelta(\pred_t^* - f(x_t))  \right\}\right] 
	\end{align*}
	Now, observe that $y^*_t$ is a function of $p_t$ and $\partial \loss(\pred^*_t,y_t)$ is a function of $y_t$ and $p_t$. Hence, we may pass to a further upper bound by inserting $\sup_{\eta_t,\mu_t}$, and by replacing each subgradient with the respective $\eta_t$ and each $\pred_t^*$ with $\mu_t$:
	\begin{align*}
		&\multiminimax{\sup_{x_t}\sup_{p_t\in\cP} \En_{y_t \sim p_t}\sup_{\eta_t\in[-G,G]}\sup_{\mu_t} \En_{\epsilon_t}}_{t=1}^n\left[\sup_{f\in\F} \left\{\sum_{t=1}^n   2 \epsilon_t \eta_t \cdot \left( \mu_t - f(x_t) \right) - \LowerDelta(\mu_t - f(x_t))  \right\}\right] \\
		&= \multiminimax{\sup_{x_t \in \X} \sup_{\eta_t \in [-G,G]} \sup_{\mu_{t}} \En_{\epsilon_t}}_{t=1}^n\left[\sup_{f\in\F} \left\{\sum_{t=1}^n   2 \epsilon_t \eta_t \cdot \left( \mu_t - f(x_t) \right) - \LowerDelta(\mu_t - f(x_t))  \right\}\right] 
	\end{align*}
	Since each $\eta_t$ range over $[-G,G]$, we can represent it as $G$ times the expectation of a random variable $u_t\in\{-1,1\}$. Denoting this distribution by $q_t$, by Jensen's inequality 
	\begin{align*}
		V_n &\leq \multiminimax{\sup_{x_t,\mu_t,q_t}\Ex_{\epsilon_t}}_{t=1}^n \left\{\sup_{f\in\F}\left[ \sum_{t=1}^{n} 2G\epsilon_t \En(u_t)(\mu_t-f(x_t)) - \LowerDelta(\mu_t-f(x_t)) \right] \right\} \\
		&\leq \multiminimax{\sup_{x_t,\mu_t,q_t}\Ex_{u_t}\Ex_{\epsilon_t}}_{t=1}^n \left\{\sup_{f\in\F}\left[ \sum_{t=1}^{n} 2G\epsilon_t u_t (f(x_t)-\mu_t) - \LowerDelta(\mu_t-f(x_t)) \right] \right\} \\
		&= \multiminimax{\sup_{x_t,\mu_t}\max_{u_t\in\{-1,1\}}\Ex_{\epsilon_t}}_{t=1}^n \left\{\sup_{f\in\F}\left[ \sum_{t=1}^{n} 2G\epsilon_t u_t (f(x_t)-\mu_t) - \LowerDelta(\mu_t-f(x_t)) \right] \right\} 
	\end{align*}
	Since for any fixed $u_t\in\{+1,-1\}$, the distribution of $\epsilon_t u_t$ is the same as that of $\epsilon_t$, the above expression is simply
	\begin{align*}
		&\multiminimax{\sup_{x_t,\mu_t}\Ex_{\epsilon_t}}_{t=1}^n \left\{\sup_{f\in\F}\left[ \sum_{t=1}^{n} 2G\epsilon_t (f(x_t)-\mu_t) - \LowerDelta(\mu_t-f(x_t)) \right] \right\} \\
		&= \sup_{\x, \bmu} \Es{\epsilon}{ \sup_{f\in\F} \left\{\sum_{t=1}^n   2G \epsilon_t \left( f(\x_t(\epsilon)-\bmu_t(\epsilon)) \right) - \LowerDelta\left(\bmu_t(\epsilon) - f(\x_t(\epsilon))\right)   \right\} }
	\end{align*}
	which is the same as the desired upper bound in \eqref{eq:val_upper_0_alpha}, in the tree notation.
\end{proof}

\begin{proof}[\textbf{Proof of Lemma~\ref{lem:genmass}}]
It holds that
\begin{align*}
		&\Es{\epsilon}{ \max_{\w\in W}\left\{ \sum_{t=1}^{n} 2C \epsilon_t \w_t(\epsilon) -      \Delta\left(\w_t(\epsilon)\right)  \right\} }\\
& = \Es{\epsilon}{\inf_{\lambda > 0} \frac{1}{\lambda} \log\left(\sum_{\w \in W} \exp\left(\lambda \left(\sum_{t=1}^{n} 2C \epsilon_t \w_t(\epsilon) -      \Delta\left(\w_t(\epsilon)\right)  \right)\right) \right)}
\end{align*}
which, by Jensen's inequality, is upper bounded by
\begin{align*}
&\inf_{\lambda > 0}\left\{  \frac{1}{\lambda} \log\left(\sum_{\w \in W} \Es{\epsilon}{\exp\left(\lambda \left(\sum_{t=1}^{n} 2C \epsilon_t \w_t(\epsilon) -      \Delta\left(\w_t(\epsilon)\right)  \right)\right)} \right) \right\}\\
& = \inf_{\lambda > 0}\left\{  \frac{1}{\lambda} \log\left(\sum_{\w \in W} \Es{\epsilon}{\prod_{t=1}^{n} \exp\left(\lambda \left(2C \epsilon_t \w_t(\epsilon) -      \Delta\left(\w_t(\epsilon)\right)   \right)\right)} \right) \right\}
\end{align*}		
Since $e^{x} + e^{-x} \le 2 e^{x^2/2}$, we have that 
\begin{align*}
\Es{\epsilon_n}{\exp\left(\lambda \left(2C \epsilon_n \w_n(\epsilon) -      \Delta\left(\w_n(\epsilon)\right)  \right)\right)} &\le \exp\left(2C^2 \lambda^2 \w_n(\epsilon)^2 -    \lambda  \Delta\left(\w_n(\epsilon)\right)  \right)\\
&= \exp\left(2C^2 \lambda^2 \w_n(\epsilon)^2 -    \lambda  \Gamma\left(\w_n(\epsilon)^2\right)  \right)
\end{align*}
By definition of conjugacy, we pass to a further upper bound of
\begin{align*}
&\exp\left( \lambda \Gamma^*\left(2C^2 \lambda \right) \right)\le  \exp\left( \lambda \Gamma^*\left(2C^2 \lambda  \right)  \right)
\end{align*}
where the last step is because $\Gamma^*$ is non-decreasing. Hence we have that 
\begin{align*}
& \frac{1}{\lambda} \log\left(\Es{\epsilon_{1:n}}{\prod_{t=1}^{n} \exp\left(\lambda \left(2C \epsilon_t \w_t(\epsilon) -      \Delta\left(\w_t(\epsilon)\right)  \right)\right)}\right) \\
&~~~~~~~~~~~ \le \frac{1}{\lambda}\log\left(\Es{\epsilon_{1:n-1}}{\prod_{t=1}^{n-1} \exp\left(\lambda \left(2C \epsilon_t \w_t(\epsilon) -      \Delta\left(\w_t(\epsilon)\right)  \right)\right)}\right) +  \Gamma^*\left(2C^2 \lambda  \right) 
\end{align*}
Proceeding in similar fashion from $n-1$ down to $1$, we arrive at an upper bound of
\begin{align*}
\frac{1}{\lambda}\log\left| W\right| + n \Gamma^*\left(2C^2 \lambda \right) 
\end{align*}
This proves the first claim. The second statement (which already appears in \cite{RakSriTew10}) is proved similarly, except the tuning value $\lambda$ is chosen at the end, and we need to account for the worst-case $\ell_2$ norm along any paths. We provide the proof here for completeness. For any tree $\w\in W$,
\begin{align*}
	\En \left[ \exp\left\{ \sum_{t=1}^{n} \lambda\epsilon_t \w_t(\epsilon) \right\} ~\middle|~ \epsilon_{1:n-1} \right] &\leq \exp\left\{ \sum_{t=1}^{n-1} \lambda\epsilon_t \w_t(\epsilon) \right\} \exp\left\{  \lambda^2 \w_n(\epsilon)^2/2\right\}  \\
	&\leq \exp\left\{ \sum_{t=1}^{n-1} \lambda\epsilon_t \w_t(\epsilon) \right\} \max_{\epsilon_n}\exp\left\{  \lambda^2 \w_n(\epsilon)^2/2\right\}
\end{align*}
Continuing in this fashion backwards to $t=1$, for any $\w\in W$
$$\En \left[ \exp\left\{ \sum_{t=1}^{n} \lambda\epsilon_t \w_t(\epsilon) \right\} \right] \leq \max_{\epsilon_1,\ldots,\epsilon_n} \exp\left\{  (\lambda^2/2) \sum_{t=1}^n \w_n(\epsilon)^2 \right\} $$
and thus
$$\En \left[ \sum_{\w\in W}\exp\left\{ \sum_{t=1}^{n} \lambda\epsilon_t \w_t(\epsilon) \right\} \right] \leq |W|\max_{\epsilon_1,\ldots,\epsilon_n}\max_{\w\in W} \exp\left\{  (\lambda^2/2) \sum_{t=1}^n \w_n(\epsilon)^2 \right\} \ . $$
Choosing 
$$\lambda = \sqrt{\frac{2\log |W|}{\max_{\epsilon_{1:n}, \w\in W} \sum_{t=1}^n \w_n(\epsilon)^2}}$$
we obtain
\begin{align*}
	\En \max_{\w\in W}\left[ \sum_{t=1}^{n} \epsilon_t \w_t(\epsilon) \right] 
	&\leq \frac{1}{\lambda}\log \En \left[ \sum_{\w\in W}\exp\left\{ \sum_{t=1}^{n} \lambda\epsilon_t \w_t(\epsilon) \right\} \right] \\
	&\leq \sqrt{2\log|W| \cdot \max_{\w\in W, \epsilon_{1:n}} \sum_{t=1}^n \w_n(\epsilon)^2}
\end{align*}
\end{proof}

\begin{proof}[\textbf{Proof of Lemma~\ref{lem:general_decomp}}]
Fix $\gamma>0$. Let $V'$ be a sequential $\gamma/2$-cover of $\G$ on $\z$ in the $\ell_\infty$ sense, i.e.
$$\forall \epsilon,~~ \forall g\in\G,~~ \exists \v'\in V' \mbox{~~~s.t.~~~} \max_{t \in [n]} \left|g(\z_t(\epsilon))-\v'_t(\epsilon)\right| \leq \gamma/2$$ 
We now modify $V'$ to construct a $\gamma$-cover of $\G$ on $\z$, which we shall denote by $V$. The $\gamma$-cover is built as follows. For every $\v' \in V'$ we include $\v$ in $V$ as defined by a soft-thresholding operation:
$$
\forall \epsilon \in \{\pm1\}^n, \forall t \in [n], ~~~ \v_t(\epsilon) = \left\{\begin{array}{cl}
0 & \textrm{ if } |\v'_t(\epsilon)| \le \gamma/2\\
\sign(\v'_t(\epsilon)) \left(|\v'_t(\epsilon)|  - \gamma/2 \right) & \textrm{ otherwise}
\end{array} \right.
$$
Since we change each $\v' \in V'$ only by $\gamma/2$ on each coordinate, $V$ is indeed a $\gamma$-cover in the $\ell_\infty$ sense. Also note that by the way we constructed $V$ from $V'$, we also have that for every $\epsilon$ and any $g \in \G$, there exists a $\v \in V$ that is $\gamma$-close in the $\ell_\infty$ sense and for this $\v$, $|g(\z_t(\epsilon))| \geq |\v_t(\epsilon)|$ for every $t$. Hence, $\Delta(g(\z_t(\epsilon))) \ge \Delta(\v_t(\epsilon))$ by the assumption that $\Delta$ is nondecreasing on $\reals_{\geq 0}$ and non-increasing on $\reals_{\leq 0}$. Denote such a $\gamma$-close tree $\v$ by $\v[\epsilon,g]$ to make the dependence on $g,\epsilon$ explicit. Since, for all $\epsilon$ and all $t$, $\Delta(g(\z_t(\epsilon))) \ge \Delta(\v[\epsilon,g]_t(\epsilon))$ we have,
\begin{align}
	\label{eq:dudley_initial_split}
	&\En \sup_{g\in\G}\left[ \sum_{t=1}^{n} 2 C \epsilon_t g(\z_t(\epsilon)) - \Delta\left(g(\z_t(\epsilon))\right)  \right] \\
	&=\En \sup_{g\in\G}\left[ \sum_{t=1}^{n} 2 C \epsilon_t  \Big(g(\z_t(\epsilon))-\v[\epsilon,g]_t(\epsilon)\Big)   +2C\epsilon_t\v[\epsilon,g]_t(\epsilon) -  \Delta(g(\z_t(\epsilon))) \right]  \notag\\
	&\le \En \sup_{g\in\G}\left[ \sum_{t=1}^{n} 2C \epsilon_t  \Big(g(\z_t(\epsilon))-\v[\epsilon,g]_t(\epsilon)\Big)     +2C\epsilon_t\v[\epsilon,g]_t(\epsilon) -  \Delta(\v[\epsilon,g]_t(\epsilon))  \right]  \notag
\end{align}
Since $\v[\epsilon,g]$ ranges over $V$, the last expression is upper bounded by  
\begin{align}
	\label{eq:decomp_chaining_1}
	\En \sup_{g\in\G}\left[ \sum_{t=1}^{n} 2C \epsilon_t \Big(g(\z_t(\epsilon))-\v[\epsilon,g]_t(\epsilon)\Big)   \right]  + \En \max_{\v\in V}\left[\sum_{t=1}^n 2C \epsilon_t\v_t(\epsilon) -  \Delta(\v_t(\epsilon)) \right] 
\end{align}
Now let $\v[\epsilon,g]$ be denoted by $\v^0[\epsilon,g]$ and $V$ be denoted by $V^0$. Let $\beta_j=2^{-j}\gamma$ and let $V^j$ denote a sequential $\beta_j$-cover of $\G$ on the tree $\z$, for $j=1,\ldots,N$, $N\geq 1$ to be specified later. We can now write the first term in \eqref{eq:decomp_chaining_1} as the constant $C$ times 
\begin{align*}
		&\En \sup_{g\in\G}\left[ \sum_{t=1}^{n} 2 \epsilon_t \Big(g(\z_t(\epsilon))-\v^N[\epsilon,g]_t(\epsilon)\Big) + \sum_{t=1}^{n} \sum_{j=1}^N 2\epsilon_t \Big(\v^{j}[\epsilon,g]_t(\epsilon)-\v^{j-1}[\epsilon,g]_t(\epsilon)\Big)   \right]  \\
			&\le \En \sup_{g\in\G}\left[ \sum_{t=1}^{n} 2\epsilon_t \Big(g(\z_t(\epsilon))-\v^N[\epsilon,g]_t(\epsilon)\Big)\right] + \sum_{j=1}^N \En \sup_{g\in\G}\left[ \sum_{t=1}^{n}  2\epsilon_t \Big(\v^{j}[\epsilon,g]_t(\epsilon)-\v^{j-1}[\epsilon,g]_t(\epsilon)\Big)   \right]  
\end{align*}
Observe that $|g(\z_t(\epsilon))-\v^N[\epsilon,g]_t(\epsilon)| \le 2 \beta_N$, and hence the first term above is upper-bounded by $4 \beta_N n$. 

We now upper bound the second term. Fix $\rho\in(0,\gamma)$ and choose $N=\max\{j:\beta_j>2\rho\}$. Then $\beta_{N+1}\leq 2\rho$ and $\beta_N \leq 4\rho$. Further, $\beta_{N+1}>\alpha$. Then second term is upper bounded via Lemma \ref{lem:genmass} by
\begin{align*}
	\sum_{j=1}^N 3\beta_j\sqrt{2 n\ \log(|V^j|\ |V^{j-1}|)} \leq 12 \sqrt{n}\int_{\rho}^\gamma \sqrt{\log\cN_\infty(\delta, \G,\z)}d\delta
\end{align*}
and, finally, the second term in \eqref{eq:decomp_chaining_1} is upper bounded via Lemma \ref{lem:genmass} by
$$\inf_{\lambda>0}\left\{\frac{1}{\lambda}\log \mathcal{N}_\infty\left(\tfrac{\gamma}{2},\G,\z\right) +  n\ \Gamma^*\left(2 C^2 \lambda  \right)\right\}  $$
Combining the results,
\begin{align*}
	&\En \sup_{g\in\G}\left[ \sum_{t=1}^{n} 2 C \epsilon_t g(\z_t(\epsilon)) - \Delta\left(g(\z_t(\epsilon))\right) \right]  \\
	&\leq  C\inf_{\rho\in(0,\gamma)} \left\{ 4 \rho n + 12 \sqrt{n}\int_{\rho}^\gamma \sqrt{\log\cN_\infty(\delta, \G,\z)}d\delta\right\} + \inf_{\lambda>0}\left\{\frac{1}{\lambda}\log \mathcal{N}_\infty\left(\tfrac{\gamma}{2},\G,\z\right) +  n\  \Gamma^*\left(2 C^2 \lambda  \right) \right\}  
\end{align*}
Since $\gamma$ was chosen arbitrarily, the result follows.

\end{proof}

\begin{proof}[\textbf{Proof of Lemma~\ref{lem:lower_bound_two_points}}]
	Recall that by definition,
	\begin{align*}
	\loss(y^*,y_t) - \loss(f(x_t),y_t) &= \partial \loss(y^*,y_t) \cdot \left( y^* - f(x_t) \right) - \Delta_{y^*,f(x_t)}^{~y_t}.
	\end{align*}
	From Eq.~\eqref{eq:valbnd0} in the proof of Lemma~\ref{lem:value_upper_bound}, 
	\begin{align*}
		V_n &= \multiminimax{\sup_{x_t}\sup_{p_t\in\cP} \En_{y_t}}_{t=1}^n\left[\sup_{f\in\F} \left\{\sum_{t=1}^n   \loss(\pred^*_t,y_t) -  \loss(f(x_t),y_t) \right\}\right] \\
		&= \multiminimax{\sup_{x_t}\sup_{p_t\in\cP} \En_{y_t}}_{t=1}^n\left[\sup_{f\in\F} \left\{\sum_{t=1}^n \partial \loss(\pred^*_t,y_t) \cdot \left(\pred^*_t - f(x_t) \right) - \Delta_{\pred^*_t,f(x_t)}^{~y_t} \right\}\right]
	\end{align*}
	The above inequality holds true if $\pred^*_t$ ensures $\Es{y \sim p_t}{\partial \loss(\pred^*_t,y)} = 0$. Let us now pass to a lower bound by restricting the set of possible optima $\pred^*_t$ to be in $S$ and the set of associated distributions to be two-point uniform distributions on the corresponding $y_1(\pred^*_t),y_2(\pred^*_t)$. Recall that by definition  
	$$\UpperDelta_{S}(f(x_t)-s) \geq \max\left\{\Delta_{s,f(x_t)}^{~y_1}, \Delta_{s,f(x_t)}^{~y_2}\right\} $$
	The lower bound is then
	\begin{align*}
		V_n & \ge R \multiminimax{\sup_{x_t} \sup_{s_t\in S} \En_{\epsilon_t}}_{t=1}^n\left[\sup_{f\in\F} \left\{\sum_{t=1}^n \epsilon_t \cdot \left(s_t - f(x_t) \right) - \frac{1}{R} \UpperDelta_{S}(f(x_t)-s_t) \right\}\right]\\
		&\geq R \sup_{\x} \Es{\epsilon}{\sup_{f\in\F} \left\{\sum_{t=1}^n \epsilon_t \cdot \left(f(\x_t(\epsilon)-\bmu_t(\epsilon)) \right) - \frac{1}{R}\UpperDelta_{\bmu_t(\epsilon)}(f(\x_t(\epsilon))-\bmu_t(\epsilon)) \right\}}
	\end{align*}
	for any $S$-valued tree $\bmu$.
\end{proof}

\begin{proof}[\textbf{Proof of Lemma~\ref{lem:lower_p_greater_2}}]
	Fix a $\beta>0$, and set $n=\fat_\beta(\F)$. Suppose $\x$ is an $\X$-valued tree of depth $n$ that is $\beta$-shattered by $\F$:
	$$\forall \epsilon, \exists f^\epsilon \in\F ~~~\mbox{s.t.}~~~ \epsilon_t (f^\epsilon(\x_t(\epsilon))-\bmu_t(\epsilon))\geq \beta/2$$
	where $\bmu$ is the witness to shattering. 
	Then from \eqref{eq:lower_bd_rad_general} with the particular choices of $\x$ and $\bmu$ described above,
	\begin{align}
		V_n &\geq \En \sup_{f\in\F}\left[ \sum_{t=1}^{n} R\epsilon_t (f(\x_t(\epsilon))-\boldsymbol{\mu}_t(\epsilon)) - \UpperDelta_{\bmu_t(\epsilon)}(f(\x_t(\epsilon))-\bmu_t(\epsilon))  \right] \\
		&\geq \En \sup_{f\in\F}\left[ \sum_{t=1}^{n} R\epsilon_t (f(\x_t(\epsilon))-\boldsymbol{\mu}_t(\epsilon)) - \frac{R}{2}|f(\x_t(\epsilon))-\boldsymbol{\mu}_t(\epsilon)|  \right] \\
		&\geq \En \left[ \sum_{t=1}^{n} R\epsilon_t (f^\epsilon(\x_t(\epsilon))-\boldsymbol{\mu}_t(\epsilon)) - \frac{R}{2}|f^\epsilon(\x_t(\epsilon))-\boldsymbol{\mu}_t(\epsilon)|  \right] 
	\end{align}
	Using the definition of shattering, we can further lower bound the above quantity by
	\begin{align*}
		&\En \left[ \sum_{t=1}^{n} \frac{R}{2}|f^\epsilon(\x_t(\epsilon))-\boldsymbol{\mu}_t(\epsilon)|  \right] \geq 
		\frac{Rn\beta}{2}.
	\end{align*}
	Now, suppose $\fat_\beta(\F)=1/\beta^p$, $p>0$. Then $n=\fat_\beta(\F)$ implies $\beta = n^{-1/p}$. The result follows.
\end{proof}

\begin{proof}[\textbf{Proof of Lemma~\ref{lem:lower_p_less}}]
Assume that $d=\fat_\beta(\F') \leq n$. Let $\z$ be an $\X$-valued tree of depth $d$ that is $\beta$-shattered by $\F'$ with a witness tree $\s$. Observe that the functions $f^\epsilon$ that guarantee 
\begin{align}
	\label{eq:fat_def}
	\forall t\in[n], ~\epsilon_t(f^\epsilon(\z_t(\epsilon))-\s_t(\epsilon))\geq \beta/2
\end{align}
do not necessarily take on values close to the $\s_t(\epsilon)\pm \beta/2$ interval. We augment $\F'$ with $2^d$ functions $g^\epsilon$ that take on the same values as $f^\epsilon$, except on points on the $\z$ tree where, for some choice $\kappa$, we have,
$$
g^\epsilon(\z_t(\epsilon)) = \epsilon_t \beta/2  + \kappa~.
$$
Let $\F$ be the resulting class of functions, and $\G=\F\setminus\F'$. We now argue that $\fat_\beta(\F)$ cannot be more than $2d+4$, as we have only added at most $2^{d}$ functions to $\F'$. Suppose for the sake of contradiction that there exists a tree $\z$ of depth at least $2d+5$ shattered by $\F$. There must exist $2^{2d+5}$ functions that shatter $\z$ and only at most $2^{d}$ of them can be from $\G$. Let us label the leaves of $\z$ with the functions that shatter the corresponding path from the root; these functions are clearly distinct. Order the leaves of the tree in any way, and observe that there must exist a pair of functions from $\G$ with indices differing by at least $2^{d+4}$. It is easy to see that such two leaves can only have a common parent at $d+3$ levels from the leaves, and this yields a complete binary subtree of size $d+1$ that is shattered by functions in $\F'$, a contradiction.

We will now use the function class $\F$ to prove a lower bound. Recall that $\z$ is an $\X$-valued tree of depth $\fat_\beta$ that is $\beta$-shattered by $\G\subseteq\F$, 
with a witness tree having the constant $\kappa$ at every node. We will now show a construction of particular trees of depth 
\begin{align}
	\label{eq:larger_n}
	n' = \left\lceil \frac{n}{\fat_\beta} \right\rceil \fat_\beta
\end{align}
using the tree $\z$.  Define $k = \lceil \frac{n}{\fat_\beta} \rceil = \frac{n'}{\fat_\beta}\geq 1$ and consider the $\X$-valued tree $\x$ and the $\reals$-valued tree $\boldsymbol{\mu}$ of depth $n'$ constructed as follows. For any path $\epsilon \in \{\pm 1\}^{n'}$ and any $t \in [n']$, set 
$$
\x_t(\epsilon) = \z_{\lceil \frac{t}{k}\rceil}\left(\tilde{\epsilon} \right), ~~~ \boldsymbol{\mu}_t(\epsilon) = \kappa
$$
where $\tilde{\epsilon} \in \{\pm 1\}^{\fat_\beta}$  is the sequence of signs specified as 
$$
\tilde{\epsilon} = \left( \sign\left(\sum_{j=1}^k \epsilon_j\right) ,\sign\left(\sum_{j=k+1}^{2k} \epsilon_j\right), \ldots, \sign\left(\sum_{j= k \left(\fat_\beta - 1\right)}^{k\, \fat_\beta} \epsilon_j\right) \right) .
$$
We now lower bound \eqref{eq:lower_bd_rad_general} by choosing the particular $\x,\boldsymbol{\mu}$ defined above:
\begin{align*}
V_{n'} &\geq R\ \En \sup_{f\in\F}\left[ \sum_{t=1}^{n'} \epsilon_t (f(\x_t(\epsilon))-\kappa) - \frac{1}{R}\bDelta_{\kappa}(f(\x_t(\epsilon))-\kappa)  \right]\\
&= R\ \En \sup_{f\in\F}\left[ \sum_{t=1}^{n'} \epsilon_t (f(\z_{\lceil \frac{t}{k}  \rceil}(\tilde{\epsilon}))- \kappa ) - \frac{1}{R} \bDelta_{\kappa}(f(\z_{\lceil \frac{t}{k}  \rceil}(\tilde{\epsilon}))-\kappa)  \right] \ .
\end{align*}
Splitting the sum over $t$ into $\fat_\beta$ blocks, the above expression is equal to
\begin{align*}
&R\ \En \sup_{f\in\F}\left[ \sum_{i=1}^{\fat_\beta}\sum_{j=(i-1)k + 1}^{i \cdot k} \epsilon_j (f(\z_i(\tilde{\epsilon}))- \kappa) - \frac{1}{R} \bDelta_\kappa(f(\z_i(\tilde{\epsilon}))-\kappa)  \right] \\
&=R\ \En \sup_{f\in\F}\left[ \sum_{i=1}^{\fat_\beta} (f(\z_i(\tilde{\epsilon}))-\kappa) \left(\sum_{j=(i-1)k + 1}^{i \cdot k} \epsilon_j \right) - \frac{k}{R}\ \bDelta_\kappa(f(\z_i(\tilde{\epsilon}))-\kappa)  \right] \\
&=R \ \En \sup_{f\in\F}\left[ \sum_{i=1}^{\fat_\beta} \tilde{\epsilon}_i(f(\z_i(\tilde{\epsilon}))-\kappa) \left|\sum_{j=(i-1)k + 1}^{i \cdot k} \epsilon_j\right|  - \frac{k}{R}\ \bDelta_\kappa(f(\z_i(\tilde{\epsilon}))-\kappa)  \right]
\end{align*}
where the last step follows by the definition of $\tilde{\epsilon}$. Recall that $\z$ is shattered by the subset $\G$ and that the functions in $\G$ stay close to the witness tree $\s$. 
We obtain a lower bound
\begin{align*}
	R \ \En \sup_{g\in\G}\left[ \sum_{i=1}^{\fat_\beta} \tilde{\epsilon}_i(g(\z_i(\tilde{\epsilon}))-\kappa) \left|\sum_{j=(i-1)k + 1}^{i \cdot k} \epsilon_j\right|  - \frac{k}{R}\ \bDelta_\kappa(g(\z_i(\tilde{\epsilon}))-\kappa)  \right]  &\geq R \ \En \sum_{i=1}^{\fat_\beta} \left( \frac{\beta}{2} \left|\sum_{j=(i-1)k + 1}^{i \cdot k} \epsilon_j\right|  - \frac{k}{R} \ \bDelta_\kappa\left(\frac{\beta}{2}  \right)\right) \\
	&\geq R\ \fat_\beta(\F) \left( \frac{\beta}{2} \sqrt{\frac{k}{2}} -  \frac{k}{R} \ \bDelta_\kappa\left(\frac{\beta}{2}  \right) \right)
\end{align*}
where we used Khinchine's inequality in the last step. By the definition of $k$,
\begin{align*}
	\fat_\beta(\F)\frac{\beta}{2} \sqrt{\frac{k}{2}} 
 =  \fat_\beta(\F)\frac{\beta}{2} \sqrt{\frac{n'}{2\, \fat_\beta(\F)}} =  \frac{1}{2 \sqrt{2}} \beta \sqrt{n' \fat_\beta(\F)} 
\end{align*}
and $\fat_\beta(\F) k = n'$ and so we conclude that,
\begin{align*}
	V_{n'} \geq    \frac{R\ \beta}{2} \sqrt{\frac{n' \fat_\beta(\F) }{2}} -  n' \ \bDelta_\kappa\left(\frac{\beta}{2}  \right) 
\end{align*}
Since we are free to choose $\kappa$ and $\beta$, 
\begin{align}
	\label{eq:lb1}
	V_{n'} \geq    \sup_{\beta, \kappa} \left\{\frac{R\ \beta}{2} \sqrt{\frac{n' \fat_\beta(\F) }{2}} -  n' \ \bDelta_\kappa\left(\frac{\beta}{2}  \right) \right\}
\end{align}
Examining \eqref{eq:unrolled_minimax_with_swap}, we see that $V_{n}$ is nondecreasing with $n$. To see this, let $n'>n$. For $t\in\{n+1,\ldots, n'\}$, we may choose $p_t$ in \eqref{eq:unrolled_minimax_with_swap} as a delta distribution on $f^*(x_t)$, for any sequence of $x_t$, where $f^*$ is an optimal function over steps $\{1,\ldots, n\}$. Clearly, $V_{n'}\geq V_{n}$. 
In view of \eqref{eq:larger_n} and the above discussion, $V_{n'}\leq V_{2n-1}$, and thus
$$V_{2n}\geq V_{2n-1} \geq V_{n'}.$$
\end{proof}

\begin{proof}[\textbf{Proof of Theorem~\ref{thm:p_and_P}}]
		We have $\Delta(t) = K t^r$ for $r \ge 2$. It will suffice to take the conjugate of $t\mapsto K t^{r/2}$ over all of $\reals$. A straightforward calculation shows that
	$$
	\Gamma^*(s) \leq \frac{K}{2} \left(r - 2\right) \left(\frac{2s}{K r}\right)^{\frac{r}{r-2}} \leq \frac{r-2}{2e} \frac{s^{\frac{r}{r-2}}}{K^{ \frac{2}{r-2}}} 
	$$

	Combining Lemma~\ref{lem:value_upper_bound} and Lemma~\ref{lem:general_decomp}, the minimax value $V_n$ is upper bounded by
\begin{align}
	\label{eq:upper_ent_1}
\inf_{\gamma > 0}\left\{ G \inf_{\rho\in(0,\gamma)} \left\{ 4  \rho n + 12 \sqrt{n}\int_{\rho}^\gamma \sqrt{\log\cN_\infty(\delta, \F,n)}d\delta\right\}   + \inf_{\lambda>0}\left\{\frac{1}{\lambda}\log \mathcal{N}_\infty\left(\tfrac{\gamma}{2},\F,n\right) +  n\ \Gamma^*\left(2\lambda G^2   \right)\right\} \right\} 
\end{align}
Consider the case $p\in(0,2)$. By the assumption on the growth of the covering numbers, and taking $\rho=1/n$,
\begin{align*}
\int_{\rho}^\gamma \sqrt{\log\cN_\infty(\delta, \F,n)} d\delta \leq \int_{\rho}^\gamma \delta^{-p/2} \sqrt{\log(n/\delta)} d\delta \leq \frac{c\sqrt{\log n}}{2-p} \gamma^{1-p/2} 
\end{align*}
Then \eqref{eq:upper_ent_1} is upper bounded by
\begin{align*}
&\inf_{\gamma > 0}\left\{ 4G + G\sqrt{n}\frac{c\sqrt{\log n}}{2-p} \gamma^{1-p/2}    + \inf_{\lambda>0}\left\{\frac{1}{\lambda}\gamma^{-p}\log(n/\gamma) +  n\ \Gamma^*\left(2\lambda G^2 \right)\right\} \right\} 
\end{align*}
We take $\gamma\geq n^{-c}$ and divide through by $\log n$:
\begin{align*}
\frac{V_n}{n\log n} &\leq \frac{4G}{n} + \inf_{\gamma \geq n^{-c}}\left\{ c_p G n^{-1/2} \gamma^{1-p/2}    + \inf_{\lambda>0}\left\{\frac{1}{n \lambda}\gamma^{-p} +  \Gamma^*\left(2\lambda G^2 \right)\right\} \right\} 
\end{align*}
where $c_p$ is a constant that depends on $p$. Balancing the terms in the inner infimum,
\begin{align*}
\inf_{\lambda>0}\left\{\frac{1}{n \lambda}\gamma^{-p} +  \Gamma^*\left(2\lambda G^2 \right)\right\} = 
\inf_{\lambda>0}\left\{\frac{1}{n \lambda}\gamma^{-p} +  c_r \frac{(\lambda G^2)^{\frac{r}{r-2}}}{K^{ \frac{2}{r-2}}} \right\}  \leq c_{r} n^{-\frac{r}{2(r-1)}} \gamma^{-\frac{rp}{2(r-1)}} G^{\frac{r}{r-1}} K^{-\frac{1}{r-1}}
\end{align*}
where $c_r$ is a constant that depends on $r$ and may change from one expression to next. The value of $\gamma$ that balances 
$$c_p G n^{-1/2} \gamma^{1-p/2}= c_{r} n^{-\frac{r}{2(r-1)}} \gamma^{-\frac{rp}{2(r-1)}} G^{\frac{r}{r-1}}$$
is $$\gamma = c_{r,p} n^{-\frac{1}{2(r-1)+p}} G^{\frac{2}{2(r-1)+p}} K^{-\frac{2}{2(r-1)+p}}$$
and it gives

\begin{align*}
	\frac{V_n}{n\log n} &\leq c_{r,p}~ 
 \left( n^{-\frac{1}{2(r-1)+p}} G^{\frac{2}{2(r-1)+p}} K^{-\frac{2}{2(r-1)+p}} \right)^{1-p/2} G n^{-1/2}\\
& =  c_{r,p}~ n^{-\frac{r}{2(r -1) + p}} G^{\frac{2r}{2(r-1)+p}}K^{-\frac{2 - p}{2(r -1) + p}} 
\end{align*}
On the other hand, using \cite[Theorem 8]{RakSriTew14jmlr}, we have
$$
V_n  \leq cG\Rad_n(\F) 
$$
In turn, sequential Rademacher complexity $\Rad_n(\F)$ is upper bounded via \eqref{eq:rad_by_dudley} by either $\mathcal{O}(n^{1-1/p})$ or $\mathcal{O}(n^{1/2})$ for $p>2$ and $p\in(0,2)$, respectively. More precisely, for the $p>2$ regime, taking $\rho=n^{-1/p}$ we obtain
\begin{align}
	\frac{1}{n}\Rad_n(\F) &\leq 4n^{-1/p} + 12n^{-1/2}\int_{n^{-1/p}}^\infty \sqrt{\delta^{-p}\log(n/\delta)}d\delta  \\
	&= cn^{-1/p} + c\sqrt{\log(n)/n}\left[ \left(\frac{2}{2-p}\right)\delta^{(2-p)/2}\right]_{n^{-1/p}}^\infty \\
	&\leq c_p n^{-1/p} \sqrt{\log(n)}.
\end{align}
The same calculation for $p\in(0,2)$ gives $\frac{1}{n}\Rad_n(\F)\leq c_\F n^{-1/2}\sqrt{\log(n)}$.
\end{proof}

\begin{proof}[\textbf{Proof of Theorem~\ref{thm:lower_p_P}}]
From Lemma~\ref{lem:lower_p_less},
\begin{align*}
V_{n} & \geq \sup_{\beta \le 1} \left\{\frac{R \beta}{2} \sqrt{\frac{n \fat_\beta(\F) }{2}} -  n \ \bDelta_\kappa\left(\frac{\beta}{2}  \right) \right\} \ge \sup_{\beta \le 1} \left\{\frac{R\sqrt{n} \beta^{1 - p/2}}{2 \sqrt{2}}  - n K \beta^r \right\}
\end{align*}
Using $\beta = \min\left\{1,  \left(\frac{R^2}{c_{p,r} K^2 n}\right)^{\frac{1}{2r + p - 2}}\right\}$, we find that
\begin{align*}
\frac{1}{n}V_n & \geq 
c_{p,r} \min\left\{\frac{R}{\sqrt{n}} ~,~~  K^{-\frac{2-p}{ 2(r-1) + p}} R^{\frac{2r}{2(r-1)+p}} n^{-\frac{r}{2(r -1) + p }}\right\}
\end{align*}
for some constant $c_{p,r}$.

\end{proof}

\begin{proof}[\textbf{Proof of Corollary~\ref{cor:lqloss_1_2}}]
	The second derivative of this loss with respect to the first argument is given by $q (q - 1) |y - \pred|^{q-2}$, and it is lower bounded by $\frac{q (q-1)}{2}$ because $q \in (1,2)$ and $y , \pred \in [-1,1]$. This means that the loss is $q(q-1)/2$ strongly convex and so 
	$$
	\LowerDelta(x) \ge \frac{q(q-1)}{2} x^2.
	$$
	We now turn to upper bounding $\UpperDelta$. Choose $S=\{0\}$ and take $y_1=1,y_2=-1$. By symmetry of the loss function, the optimal $\pred^* = 0$, verifying property \eqref{property_two_point}. Then 
	\begin{align}
			\UpperDelta_0 (x) = \sup_{x \in \Y}\max\left\{\Delta_{0,x}^{1}, \Delta_{0,x}^{-1}\right\},
	\end{align}
	with domain $\predset-\{0\} = [-1,1]$. For any $y\in\Y$, the generalized binomial theorem gives an expansion of $\ell(\cdot,y)$ at the point $a\neq y$ as
	\begin{align*}
	\Delta^y_{a,b} &= \ell(b,y) - [\ell(a,y) + \partial_a \ell(a,y) \cdot (b-a)]  = \sum_{j=2}^{\infty} \frac{\prod_{k=0}^{j-1}(q - k)}{j!}  (a - y)^{q-j} \cdot (b - a)^j
	\end{align*}
	Then, taking $b=x$ and $a=0$,
	\begin{align*}
	\UpperDelta_0(x) &\le \sum_{j=2}^{\infty} \frac{\prod_{k=0}^{j-1} \left|q-k\right|}{j!}  |x|^j
	 =  \left(\sum_{j=2}^{\infty} \frac{\prod_{k=2}^{j-1} (k-q)}{j!}  |x|^{j-2}\right) q(q-1)x^2 
	 \end{align*}
	Since $q > 1$ we can bound the above by
	\begin{align*}
	\left(\sum_{j=2}^{\infty} \frac{(j-2)!}{j!}  |x|^{j-2}\right) q(q-1)x^2 \le \left(\sum_{j=2}^{\infty} \frac{1}{j(j-1)}\right)   q(q-1)x^2 \le 2q(q-1) x^2.
	\end{align*}
	The result follow from Theorems \ref{thm:p_and_P} and \ref{thm:lower_p_P}.
	
\end{proof}

%% file: paper-plain.bbl
\begin{thebibliography}{10}

\bibitem{AbeAgaBarRak09}
J.~Abernethy, A.~Agarwal, P.~Bartlett, and A.~Rakhlin.
\newblock A stochastic view of optimal regret through minimax duality.
\newblock In {\em Proceedings of the 22nd Annual Conference on Learning
  Theory}, 2009.

\bibitem{audibert2009fast}
J.Y. Audibert.
\newblock Fast learning rates in statistical inference through aggregation.
\newblock {\em The Annals of Statistics}, 37(4):1591--1646, 2009.

\bibitem{acbg-ascola-02}
P.~Auer, N.~Cesa-Bianchi, and C.~Gentile.
\newblock Adaptive and self-confident on-line learning algorithms.
\newblock {\em Journal of Computer and System Sciences}, 64(1):48--75, 2002.

\bibitem{AzouryWarmuth01}
K.~S. Azoury and M.~K. Warmuth.
\newblock Relative loss bounds for on-line density estimation with the
  exponential family of distributions.
\newblock {\em Machine Learning}, 43(3):211--246, June 2001.

\bibitem{bach2010self}
F.~Bach.
\newblock Self-concordant analysis for logistic regression.
\newblock {\em Electronic Journal of Statistics}, 4:384--414, 2010.

\bibitem{cb-atgbaolr-99}
N.~Cesa-Bianchi.
\newblock Analysis of two gradient-based algorithms for on-line regression.
\newblock {\em Journal of Computer and System Sciences}, 59(3):392--411, 1999.

\bibitem{cbfhhsw-huea-97}
N.~Cesa-Bianchi, Y.~Freund, D.~Haussler, D.~P. Helmbold, R.~E. Schapire, and
  M.~K. Warmuth.
\newblock How to use expert advice.
\newblock {\em Journal of the ACM}, 44(3):427--485, 1997.

\bibitem{cesa1999minimax}
N.~Cesa-Bianchi and G.~Lugosi.
\newblock Minimax regret under log loss for general classes of experts.
\newblock In {\em Proceedings of the Twelfth annual conference on computational
  learning theory}, pages 12--18. ACM, 1999.

\bibitem{PLG}
N.~Cesa-Bianchi and G.~Lugosi.
\newblock {\em Prediction, Learning, and Games}.
\newblock Cambridge University Press, 2006.

\bibitem{f-pwc-91}
D.~P. Foster.
\newblock Prediction in the worst case.
\newblock {\em Annals of Statistics}, 19(2):1084--1090, 1991.

\bibitem{gerchinovitz2013sparsity}
S.~Gerchinovitz.
\newblock Sparsity regret bounds for individual sequences in online linear
  regression.
\newblock {\em Journal of Machine Learning Research}, 14:729--769, 2013.

\bibitem{gerchinovitz2013adaptive}
S.~Gerchinovitz and J.~Yu.
\newblock Adaptive and optimal online linear regression on $\ell_1$-balls.
\newblock {\em Theoretical Computer Science}, 2013.

\bibitem{GinZin84}
E.~Gin\'e and J.~Zinn.
\newblock Some limit theorems for empirical processes.
\newblock {\em Annals of Probability}, 12(4):929--989, 1984.

\bibitem{HanRakSri13}
W.~Han, A.~Rakhlin, and K.~Sridharan.
\newblock Competing with strategies.
\newblock In {\em Conference on Learning Theory}, 2013.

\bibitem{haussler1998sequential}
D.~Haussler, J.~Kivinen, and M.~Warmuth.
\newblock Sequential prediction of individual sequences under general loss
  functions.
\newblock {\em Information Theory, IEEE Transactions on}, 44(5):1906--1925,
  1998.

\bibitem{HazMeg07}
E.~Hazan and N.~Megiddo.
\newblock Online learning with prior knowledge.
\newblock In {\em Learning Theory}, volume 4539 of {\em Lecture Notes in
  Computer Science}, pages 499--513. 2007.

\bibitem{KivWar97}
J.~Kivinen and M.~K. Warmuth.
\newblock Exponentiated gradient versus gradient descent for linear predictors.
\newblock {\em Inf. Comput.}, 132(1):1--63, 1997.

\bibitem{mf-up-98}
N.~Merhav and M.~Feder.
\newblock Universal prediction.
\newblock {\em IEEE Transactions on Information Theory}, 44:2124--2147, 1998.

\bibitem{rakhlin2012relax}
A.~Rakhlin, O.~Shamir, and K.~Sridharan.
\newblock Relax and randomize: From value to algorithms.
\newblock In {\em Advances in Neural Information Processing Systems 25}, pages
  2150--2158, 2012.

\bibitem{StatNotes2012}
A.~Rakhlin and K.~Sridharan.
\newblock Statistical learning and sequential prediction, 2012.
\newblock Available at {\small
  \url{http://stat.wharton.upenn.edu/~rakhlin/courses/stat928/stat928_notes.pdf}}.

\bibitem{rakhlin2013semi}
A.~Rakhlin and K.~Sridharan.
\newblock On semi-probabilistic universal prediction.
\newblock In {\em Information Theory Workshop (ITW), 2013 IEEE}, pages 1--5.
  IEEE, 2013.

\bibitem{RakSri14a}
A.~Rakhlin and K.~Sridharan.
\newblock Online nonparametric regression.
\newblock In {\em Conference on Learning Theory}, 2014.

\bibitem{RakSri14nonparametric}
A.~Rakhlin and K.~Sridharan.
\newblock Online nonparametric regression.
\newblock In {\em Proceedings of The 27th Conference on Learning Theory},
  volume~35, page 1232–1264, 2014.

\bibitem{RakSriTew10}
A.~Rakhlin, K.~Sridharan, and A.~Tewari.
\newblock Online learning: Random averages, combinatorial parameters, and
  learnability.
\newblock {\em Advances in Neural Information Processing Systems 23}, pages
  1984--1992, 2010.

\bibitem{RakSriTew14jmlr}
A.~Rakhlin, K.~Sridharan, and A.~Tewari.
\newblock Online learning via sequential complexities.
\newblock {\em Journal of Machine Learning Research}, 2014.
\newblock To appear.

\bibitem{RakSriTew14}
A.~Rakhlin, K.~Sridharan, and A.~Tewari.
\newblock Sequential complexities and uniform martingale laws of large numbers.
\newblock {\em Probability Theory and Related Fields}, 2014.
\newblock To appear.

\bibitem{RakSriTsy14}
A.~Rakhlin, K.~Sridharan, and A.~Tsybakov.
\newblock Entropy, minimax regret and minimax risk.
\newblock In submission, 2014.

\bibitem{vovk1995game}
V.~Vovk.
\newblock A game of prediction with expert advice.
\newblock In {\em Proceedings of the eighth annual conference on Computational
  learning theory}, pages 51--60. ACM, 1995.

\bibitem{Vovk98}
V.~Vovk.
\newblock Competitive on-line linear regression.
\newblock In {\em NIPS '97: Proceedings of the 1997 conference on Advances in
  neural information processing systems 10}, pages 364--370, Cambridge, MA,
  USA, 1998. MIT Press.

\bibitem{v-cos-01}
V.~Vovk.
\newblock Competitive on-line statistics.
\newblock {\em International Statistical Review}, 69:213--248, 2001.

\bibitem{Vovk06metric}
V.~Vovk.
\newblock Metric entropy in competitive on-line prediction.
\newblock {\em CoRR}, abs/cs/0609045, 2006.

\bibitem{vovk2006hilbert}
V.~Vovk.
\newblock On-line regression competitive with reproducing kernel hilbert
  spaces.
\newblock In {\em Theory and Applications of Models of Computation}, pages
  452--463. Springer, 2006.

\bibitem{Vovk07wild}
V.~Vovk.
\newblock Competing with wild prediction rules.
\newblock {\em Machine Learning}, 69(2):193--212, 12 2007.

\end{thebibliography}
